%% file: 0-main.tex
\def\year{2022}\relax
\documentclass[letterpaper]{article} 
\usepackage{aaai22}  
\usepackage{times}  
\usepackage{helvet}  
\usepackage{courier}  
\usepackage[hyphens]{url}  
\usepackage{graphicx} 
\usepackage{natbib}  
\usepackage{caption} 
\DeclareCaptionStyle{ruled}{labelfont=normalfont,labelsep=colon,strut=off} 
\usepackage{algorithm}
\usepackage{algorithmic}
\usepackage{hyperref}
\usepackage{mathtools}
\usepackage{amsmath}
\usepackage{multirow}
\usepackage{booktabs}
\usepackage{subfig}

\usepackage{newfloat}
\usepackage{listings}
\floatstyle{ruled}
\newfloat{listing}{tb}{lst}{}
\floatname{listing}{Listing}
\title{Enhanced Exploration in Neural Feature Selection for Deep Click-Through Rate Prediction Models via Ensemble of Gating Layers}
\author{
    Lin Guan\textsuperscript{\rm 1, *},
     Xia Xiao\textsuperscript{\rm 2},
     Ming Chen\textsuperscript{\rm 2},
     Youlong Cheng\textsuperscript{\rm 2}
}
\affiliations {
    \textsuperscript{\rm 1} School of Computing \& AI,  
        Arizona State University\\
    \textsuperscript{\rm 2} ByteDance\\
    lguan9@asu.edu, x.xiaxiao@bytedance.com, ming.chen@bytedance.com,
    youlong.cheng@bytedance.com
}

\usepackage{bibentry}

\begin{document}

\maketitle

\begin{abstract}
Feature selection has been an essential step in developing industry-scale deep Click-Through Rate (CTR) prediction systems. The goal of neural feature selection (NFS) is to choose a relatively small subset of features with the best explanatory power as a means to remove redundant features and reduce computational cost. Inspired by gradient-based neural architecture search (NAS) and network pruning methods, people have tackled the NFS problem with Gating approach that inserts a set of differentiable binary gates to drop less informative features. The binary gates are optimized along with the network parameters in an efficient end-to-end manner. In this paper, we analyze the gradient-based solution from an exploration-exploitation perspective and use empirical results to show that Gating approach might suffer from insufficient exploration. To improve the exploration capacity of gradient-based solution, we propose a simple but effective ensemble learning approach, named Ensemble Gating. We choose two public datasets, namely Avazu and Criteo, to evaluate this approach. Our experiments show that, without adding any computational overhead or introducing any hyper-parameter (except the size of the ensemble), our method is able to consistently improve Gating approach and find a better subset of features on the two datasets with three different underlying deep CTR prediction models.
\end{abstract}

\input{1-intro}

\input{2-related}
\input{3-problem}

\input{4-method}
\input{5-experiment}

\input{6-conclusion}

\bibliography{aaai22}

\input{7-appendix}

\end{document}

%% file: 1-intro.tex
\section{Introduction}
\label{sec:intro}

Deep Click-Through Rate (CTR) prediction systems have shown promising performance in many industry-scale CTR prediction tasks, but this usually comes with high computational cost and high memory usage. To make deep learning models more memory efficient, people have proposed many approaches to reduce the size of the networks, such as network pruning \cite{Xiao2019AutoPruneAN, gao2020discrete}, training with resource constraint \cite{srinivas2017training}, and adjusting feature embedding sizes \cite{joglekar2019neural, liu2021learnable}. In this work, we focus on improving memory efficiency by performing neural feature selection that chooses a relatively small subset of features with good explanatory power.

\begin{figure}
\begin{center}
\centerline{\includegraphics[width=\columnwidth]{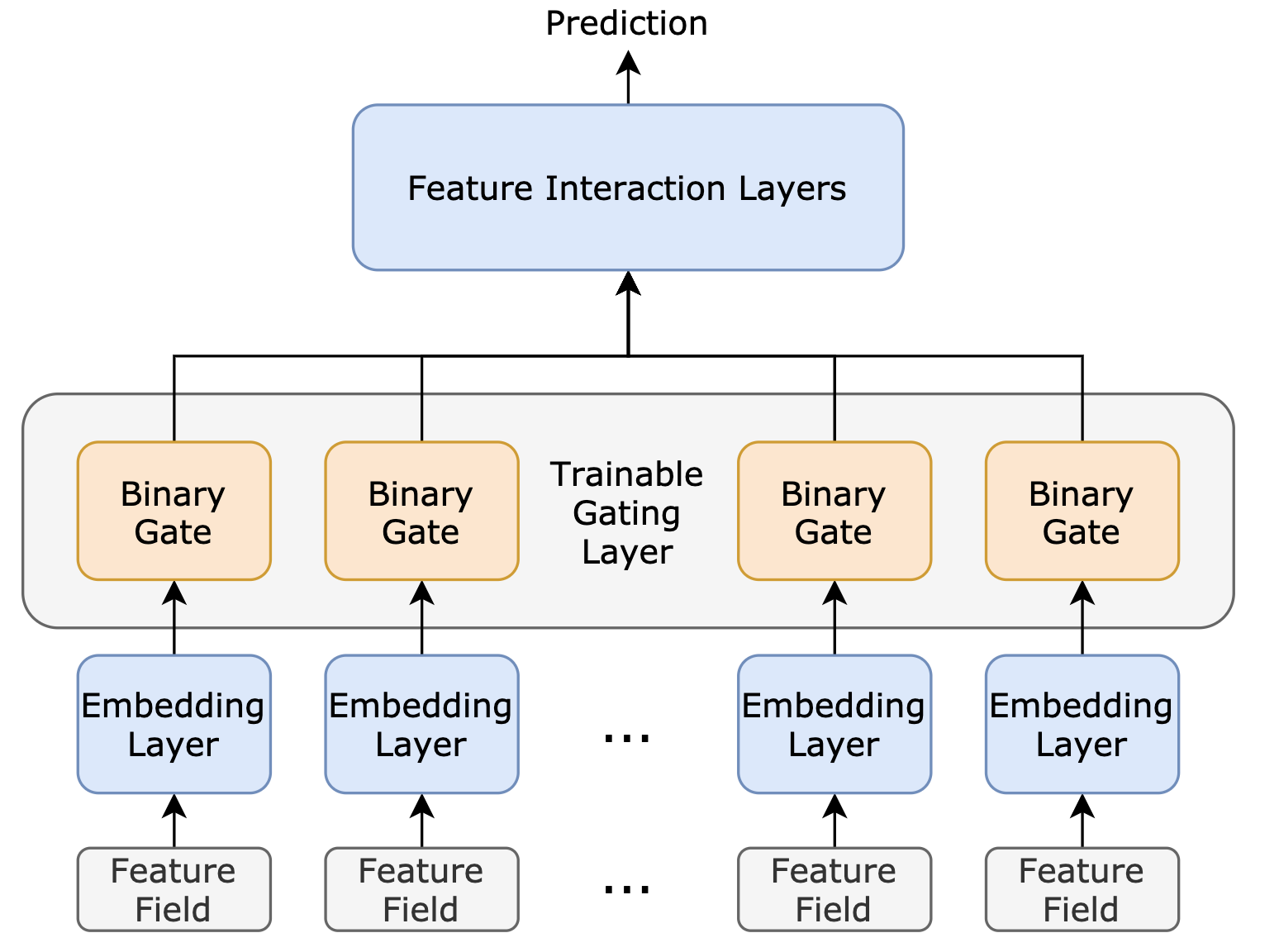}}
\caption{A typical deep learning CTR prediction model with trainable binary gates inserted between embedding layers and feature interaction layers.}
\label{fig:rec-model-arch}
\end{center}
\vspace{-0.8cm}
\end{figure}

We consider end-to-end feature selection approaches, which interpret the feature selection problem as a differentiable neural architecture search (NAS) or network pruning problem, and thus enable people to use many well-studied methods from the two relevant domains. In those approaches, to remove less predictive features, a set of ``binary" gates with trainable weights are added to the network to indicate if certain features should remain or drop (Fig. \ref{fig:rec-model-arch}). Depending on how discrete relaxation is achieved, there can be different implementations of the binary gates \cite{sheth2020differentiable, yamada2020feature}. Similar to differentiable NAS \cite{cai2018proxylessnas} and network pruning \cite{Xiao2019AutoPruneAN, gao2020discrete}, differentiable neural feature selection methods realize their functionality by iteratively performing feature pruning (updating binary gates) and network fine-tuning (updating network parameters) over a pretrained over-parameterized network. In this paper, we refer to the methods that use such a gating formula as Gating method and the gate output as gating decision.

\begin{figure}
\centering
\subfloat{\includegraphics[width=0.33\linewidth]{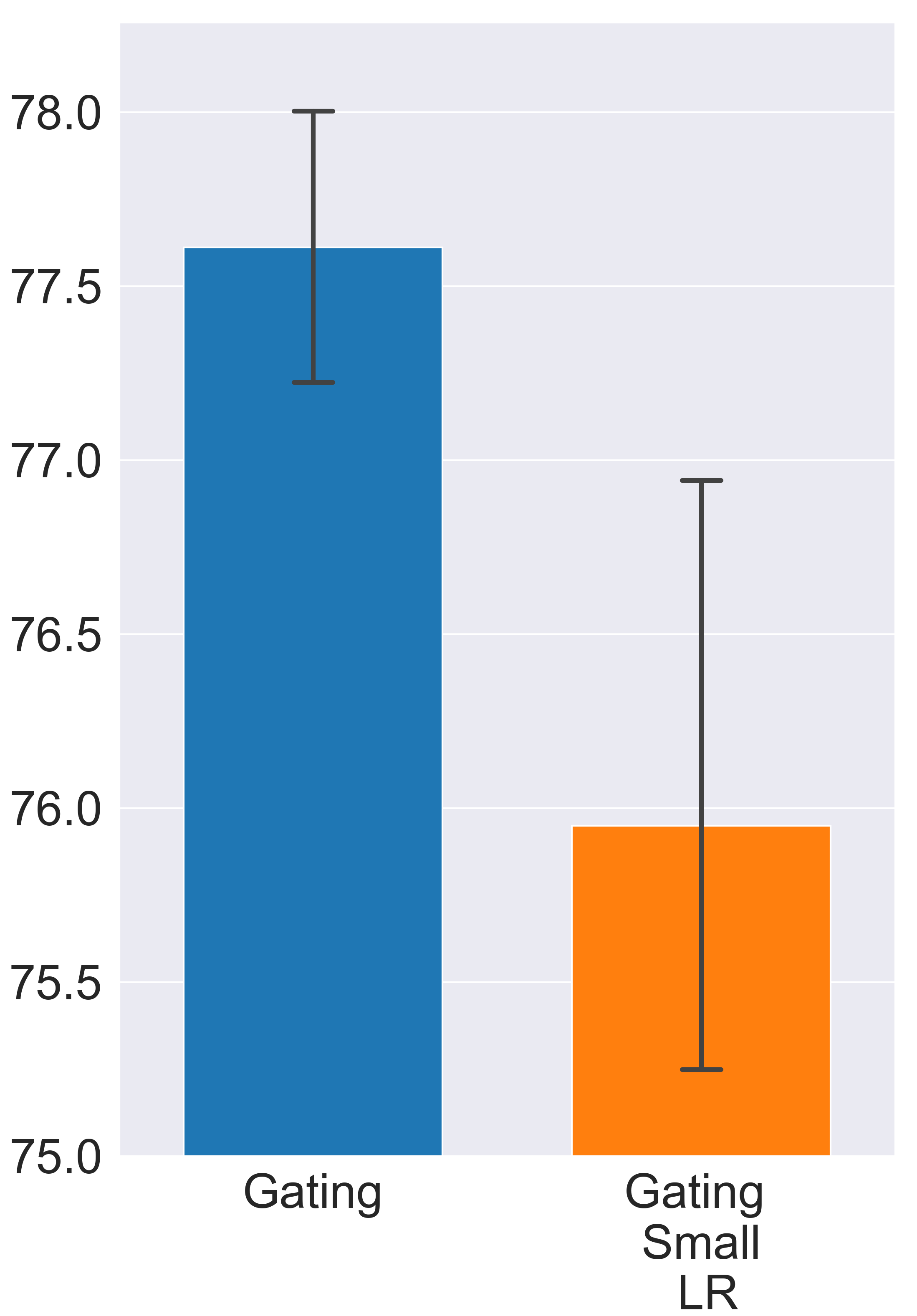}}
\subfloat{\includegraphics[width=0.33\linewidth]{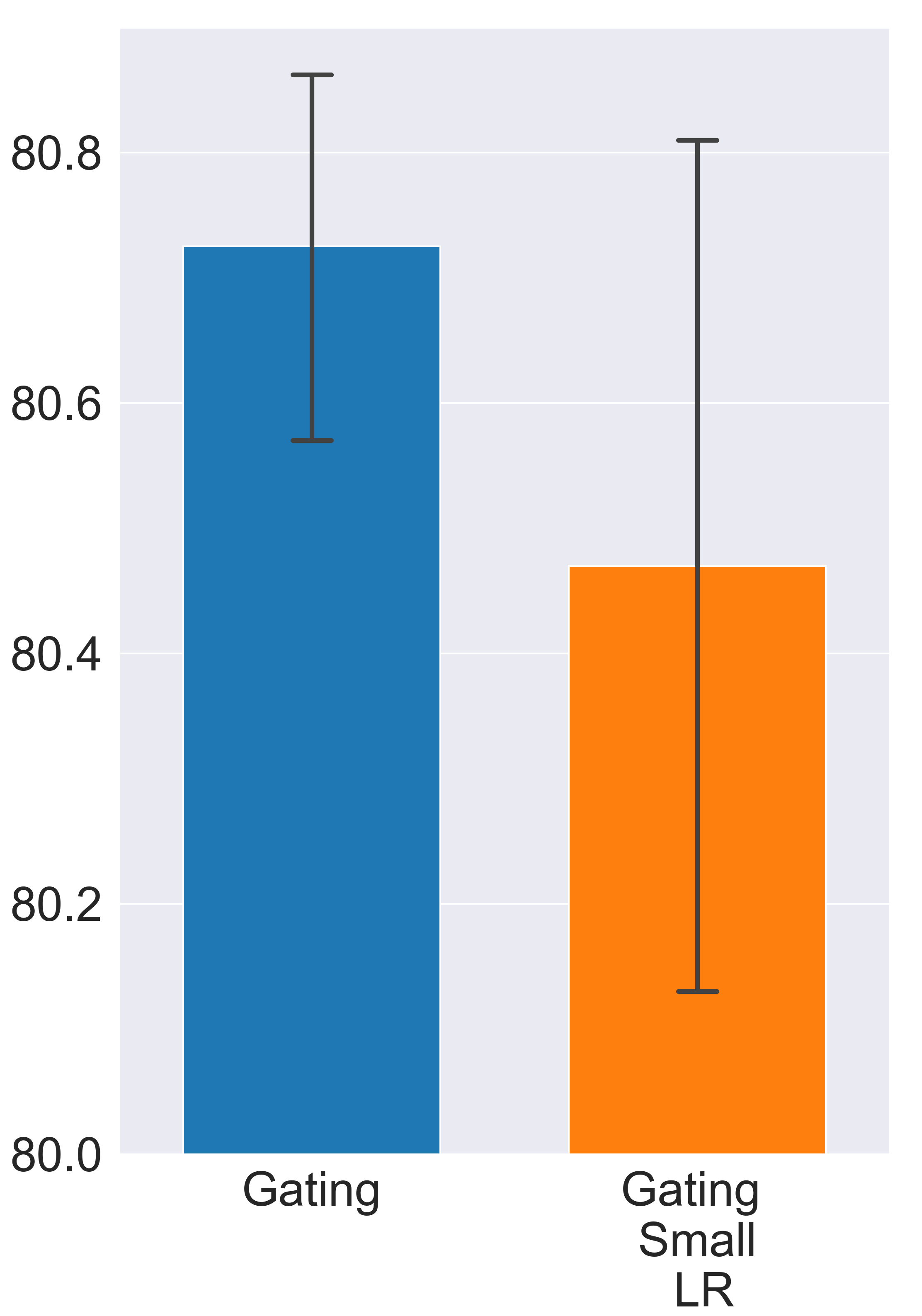}}
\caption{The testing AUC scores of DCN model that takes selected features as input. An abnormally small learning rate of the binary gates significantly degrades the performance of Gating method. Results were obtained on the Avazu dataset (left) and the Criteo dataset (right).}
\label{fig:special-expr}
\vspace{-0.2cm}
\end{figure}

Since the feature selection problem is essentially a search problem with the search space comprised of all possible gating decisions, to find the most informative subset of features, the gates must fully explore the search space by taking various gating decisions and comparing the outcomes. However, in the Gating approach, the gating exploration is mainly driven by the gradient signal, which doesn't always provide a strong enough exploration incentive, especially when the binary gates are jointly optimized with the model parameters. Here, we present a special experiment to illustrate how insufficient exploration can result in undesirable outcomes. In this experiment, we use a step function and straight through estimator \cite{Bengio2013EstimatingOP, hubara2016binarized} to implement the binary gates. Details of the implementations and the searching/training process can be found in later sections. But different from normal settings, we deliberately reduce the learning rate of binary gates by a factor of ten. The goal of this setting change is to exaggerate the negative effects of insufficient exploration. Specifically, suppose the gradient signal suggests the $i^{th}$ gate should be changed from ``on" to ``off" and suppose there is no other exploration incentive except the gradient, since the learning rate of binary gates is abnormally small, it takes more update steps for the $i^{th}$ gate to change its state. However, before the $i^{th}$ gate is turned off, the model parameters have already been fine-tuned according to current gating decision. In this case, changing gate states will instead increase the training loss. We call this phenomenon \textit{gating overfitting}. The experiment result verifies our hypothesis (Fig. \ref{fig:special-expr}). Note that gating overfitting exists even when all the hyper-parameters are properly set. 

To enhance gating exploration and prevent gating overfitting, we propose a scalable and effective ensemble learning method, named Ensemble Gating, which creates multiple groups of binary gates and randomly selects one group to perform parameters update at each step. We show that, without adding any computational overhead or introducing any hyper-parameter (except the size of the ensemble), Ensemble Gating can provide meaningful uncertainty-driven exploration. We conduct extensive experiments on two public datasets, namely Avazu and Criteo. The results demonstrate that Ensemble Gating consistently finds better subsets of features on both datasets for three different deep CTR prediction models. Further studies show that Ensemble Gating converges quickly within a few epochs, and the uncertainty-driven exploration is more effective than exploring by injecting random noise into the learning system.

%% file: 2-related.tex
\section{Related Works}

The idea of using differentiable binary gates to automatically select important features was discussed in \cite{sheth2020differentiable, yamada2020feature}. It has also been widely applied in industry and has been studied in many relevant research directions like gradient-based neural architecture search \cite{cai2018proxylessnas, liu2018darts} and network pruning \cite{han2015learning, srinivas2017training, ye2020adaptive}, which use learnable binary gates to determine whether to prune certain weight or neuron. Gating approaches in neural feature selection can be viewed as a reduced version of related network pruning approaches in the sense that binary gates are only inserted between certain layers in neural feature selection. 

Due to the discrete nature of feature selection/network pruning problems (there are only two possible actions, namely "prune" or "keep"), there is no direct way to optimize the binary gates along with the deep neural network in a fully end-to-end manner. Thus, substantial research efforts have been focusing on finding better approximation that relaxes the discrete search space to a continuous space by using some differentiable re-parameterization tricks. Works like \cite{Xiao2019AutoPruneAN} and \cite{gao2020discrete} use straight through estimator \cite{Bengio2013EstimatingOP, hubara2016binarized} to enable gradients to be propagated. Some other works also use Gumbel-Softmax \cite{jang2016categorical, maddison2016concrete} to do the discrete relaxation \cite{xie2018snas, liu2018darts}.  

The problem of neural feature selection (NFS) is also related to the neural input search (NIS) problem \cite{joglekar2019neural, ginart2019mixed}. But the latter aims to reduce memory usage by assigning varying embedding sizes to different features rather than dropping an entire feature. Many works employ a similar differentiable gating formula to tackle the NIS problem \cite{zhao2021autodim, cheng2020differentiable, liu2021learnable}, but in their frameworks, each binary gate is used to control the dimension of each embedding vector. Note that NIS and NFS are not interchangeable. Firstly, NIS can not be applied to models that require feature embeddings of the same size, such as DeepFM \cite{guo2017deepfm}. Secondly, it can be more effective to perform feature selection before applying NIS or other network pruning techniques, especially for industry-scale CTR prediction models.

Feature selection for CTR prediction systems has been explored previously \cite{ronen2013selecting, koenigstein2013xbox}. In this work, we focus on improving the gradient-based feature methods, which are shown to be more computationally efficient with large-scale deep learning models \cite{sheth2020differentiable}. Comprehensive surveys on feature selection are provided by \citeauthor{parmezan2021automatic} and \citeauthor{chandrashekar2014survey}.

%% file: 3-problem.tex
\section{Preliminaries}
\label{sec:problem}

Figure \ref{fig:rec-model-arch} depicts a typical deep learning CTR prediction model. We assume the model input $X$ consists of $\mathcal{M}$ categorical feature fields. Each raw categorical feature is initially represented by a sparse one-hot vector. Then the feature embedding layer transforms each sparsely encoded feature $x_i$ into a $V$-dimensional embedding vector $e_i$ as follows:
\small
\begin{equation}
\label{eq:emb-layer}
    e_i = E_i^{\top}x_i,
\end{equation}
\normalsize
where $E_i \in R^{D \times V}$ denotes the \textit{embedding matrix}, and $D$ denotes the size of sparse encoding. Note that in this paper, we assume a global embedding size $V$ (default value: 8) is used for all the features. Finally, all the dense feature embeddings are concatenated into an embedding matrix that is used by the remaining parts of the model (i.e. feature interaction layers) to predict the probability that a user likes an item (a.k.a. click-through rate prediction). Let $\theta$ be the parameters of feature interaction layers, and $\Theta = \{E, \theta \}$ be the set of trainable model parameters, the output of the model is given by:
\small
\begin{equation}
\label{eq:model-pred}
    \hat{y} = \phi(X|\Theta),
\end{equation}
\normalsize
where $\phi$ represents the CTR prediction model, $\hat{y}$ is the model prediction. Note that the architecture of feature interaction layers may vary in different CTR prediction models.

The CTR prediction model $\phi$ is trained to optimize the following objective:
\small
\begin{equation}
\label{eq:acc-loss}
    \min_{\Theta} \mathcal{L}(\phi(X|\Theta), \mathcal{D}),
\end{equation}
\normalsize
where $\mathcal{D}$ is the size of the training dataset and $\mathcal{L}$ is the model loss function (binary cross-entropy in usual) along with a weight decay regularization term.

%% file: 4-method.tex
\section{Method}

In this section, we first formalize the problem of neural feature selection (NFS) and discuss existing solutions that are adapted from gradient-based NAS and network pruning. Then we introduce our Ensemble Gating algorithm that aims to offer effective uncertainty-driven exploration to overcome the gating overfitting problem. 

\subsection{Neural Feature Selection} 
The goal of neural feature selection is to select $\mathcal{N}$ features from the $\mathcal{M}$ feature fields. In this paper, we formulate the NFS problem as a differentiable network pruning problem, thereby connecting it to gradient-based network pruning methods \cite{Xiao2019AutoPruneAN, ye2020adaptive} and the problem of gradient-based NAS \cite{cai2018proxylessnas}. In this formulation, an ``over-parameterized" pretrained network is provided, and our task is to remove less informative or redundant model inputs. To automate the feature selection process, a feature gating layer with trainable weights is inserted between the embedding layers and the feature interaction layers (Fig. \ref{fig:rec-model-arch}). Specifically, the feature gating layer contains $\mathcal{M}$ real-valued gating parameters $\{\alpha_i\}$, which matches the number of input feature fields. Then for each gating parameter $\alpha_i$, a differentiable \textit{binarize} function converts it into binary gate $g_i$ that determines whether to prune or keep feature embedding $e_i$:
\small
\begin{equation}
\label{eq:gating-function}
    g_i = binarize(\alpha_i) = 
    \begin{cases}
    0,& \text{if } e_i \text{ is pruned;}\\
    1,              & \text{otherwise.}
    \end{cases}
\end{equation}
\normalsize
Accordingly, the input to the feature interaction layers is replaced with the masked feature embeddings $\{\tilde{e}_i\}$:
\small
\begin{equation}
\label{eq:masked-embedding}
    \tilde{e}_i = g_i \cdot e_i
\end{equation}
\normalsize
Note that the choice of the binarize function is flexible as long as the function can (coarsely) map any real value into $\{0, 1\}$ and the backward gradient can be accurately estimated. Two commonly used binarize functions include softmax function (with properly picked temperature) and step function with straight through estimator (STE) \cite{Bengio2013EstimatingOP, hubara2016binarized}. In this paper, we are most interested in improving the exploration capacity of the STE-based solution, in which the binary gating decision is given by:
\small
\begin{equation}
\label{eq:ste-decision}
    binarize\texttt{\_}STE(\alpha_i) = 
    \begin{cases}
    1,& \text{if } \alpha_i > 0\text{;}\\
    0,              & \text{otherwise.}
    \end{cases}
\end{equation}
\normalsize
During the backward propagation, the gradient is computed as if the $binarize\texttt{\_}STE$ function were an identity function. For simplicity, we assume the Gating method is implemented with step function and STE in the remainder of this paper.

To optimize the feature selection, the gating parameters are trained to minimize the following loss:
\small
\begin{equation}
\label{eq:gating-loss}
    \min_{\alpha} \mathcal{L}(\phi(X|\Theta, \alpha), \mathcal{D}_g) + \beta_s R_{s}(\alpha),
\end{equation}
\normalsize
where $\mathcal{D}_g$ is the dataset used for gating parameters training, $R_{s}$ is the sparse regularization term that controls the degree of sparsity, and $\beta_s$ is the weight of sparse regularization. $R_{s}$ is typically defined with the difference between the target number of selected features and the number of open gates:
\begin{equation}
\small
\label{eq:sparsity-regularization}
    R_{s}(\alpha_i) = 
    \begin{cases}
    count(g_i=1),& \text{if } count(g_i=1) > target\text{;}\\
    0,              & \text{otherwise.}
    \end{cases}
\normalsize
\end{equation}
Note that we don't penalize when the number of open gates is lower than the target value, because this usually leads to an increase in training loss and the gradient will push the gating layer to reopen some gate(s) \cite{Xiao2019AutoPruneAN}.

Following previous related works, to yield the best feature selection result, the network parameters $\Theta$ and the gating parameters $\alpha$ are updated iteratively: when training network parameters, all binary gates are fixed and $\Theta$ is updated according to Eq. (\ref{eq:acc-loss}) with a mini-batch sampled from $\mathcal{D}$; when training gating parameters, the network parameters are frozen and $\alpha$ is updated according to Eq. (\ref{eq:gating-loss}) with a mini-batch sampled from $\mathcal{D}_g$. Note that it's not necessary that $\mathcal{D}$ and $\mathcal{D}_g$ are different subsets of data. In fact, our experiment results suggest that letting $\mathcal{D}$ differ from $\mathcal{D}_g$ doesn't yield better results with Gating method (details can be found in the Experiments section). Considering making $\mathcal{D} = \mathcal{D}_g$ can be more computationally efficient (because we only need to do one single backward pass at each update step when network parameters and gating parameters share the same input data), we use $\mathcal{D} = \mathcal{D}_g$ as the default setting in this paper.

\subsection{Neural Feature Selection via Ensemble of Gating Layers} 
One limitation of Gating method is that the gating decision is optimized in a fully exploitative manner, in which the only objective is to minimize the training loss by iteratively updating network parameters and gating parameters. In the Introduction section, we already show that this joint optimization doesn't provide sufficient gating exploration and can easily get trapped into sub-optimal local minima. Moreover, although the differentiable binarize function relaxes the NP-hard discrete feature selection problem to a continuous optimization problem, the problem itself is still an extremely complex search problem in essence. Hence, a better exploration strategy can definitely benefit the process of training gating parameters.  

To mitigate the gating overfitting issue and achieve more effective gating exploration, we present a simple but effective ensemble learning method called \textit{Ensemble Gating}. Rather than having only one group of binary gates as in Gating, we maintain $K$ groups of gates $\{\alpha^k_i\}^{k=1}_K$. All the groups share the same underlying network parameters $\Theta$, and for each mini-batch of training data $\mathcal{D}_{batch}$, one group of gates is randomly selected to perform parameters update:
\begin{equation}
\small
\label{eq:gating-update}
\begin{split}
     \min_{\alpha^k} \mathcal{L} & (\phi(X|\Theta, \alpha^k), \mathcal{D}_{batch}) + \beta_sR_{s}(\alpha^k), \\
     & k \sim Uniform\{1, ..., K\}.
\end{split}
\normalsize
\end{equation}
This training process is essentially the process of generating bootstrapped subsets of gating training data that is sampled with replacement from the entire dataset. Hence, the objective can also be written as:
\begin{equation}
\small
\label{eq:bootstrapped}
\begin{split}
     \min_{\alpha} \frac{1}{K} \sum^k_K (\mathcal{L} & (\phi(X|\Theta, \alpha^k), \mathcal{D}_{k}) + \beta_sR_{s}(\alpha^k)),
\end{split}
\normalsize
\end{equation}
where $\mathcal{D}_{k}$ represents the bootstrapped training samples for the $k^{th}$ group of binary gates. \\
\\
\noindent\textbf{Uncertainty-Driven Exploration } Intuitively, the exploration in Ensemble Gating is driven by the inter-group disagreements. For most important features, there can be a significant increase in training loss if they are dropped by the gates. So all the groups can soon reach a consensus on keeping those features. On the contrary, for other less informative features, the gating layer might need to take more \textit{exploratory actions} (by varying gating decisions and comparing the consequences) to determine their influence on the model performance. Randomly selecting one group to perform parameters update is actually simulating this exploration behavior. In fact, ensemble of neural networks (NNs) has been widely used as a means to estimate predictive uncertainty \cite{lakshminarayanan2016simple}. In Reinforcement Learning research, similar ensemble methods are also used to encourage more meaningful temporally-extended (deep) exploration \cite{osband2016deep, pathak2019self}. In the context of NFS, the agreement/disagreement among the ensemble of binary gates can be interpreted as the gating layer's \textit{predictive uncertainty} over feature importance. Only feature embeddings that are considered important with high confidence get fine-tuned constantly. While for features whose importance is yet to be determined, the gating layer will continuously explore different gating decisions. As a consequence, those feature embeddings get fine-tuned more equally and less frequently, thereby effectively preventing the gating overfitting issue. \\

\begin{algorithm}[tb]
\caption{Ensemble Gating}
\label{alg:algorithm-ensemble}
\textbf{Input}: Training set $\mathcal{D}$, \textit{iter}\\
\textbf{Parameter}: Pretrained model $\phi$ with parameters $\Theta$, ensemble size $K$, initialized gating parameters $\{\alpha\}_K$\\
\textbf{Output}: Final gate states $\{g\}$ 
\begin{algorithmic}[1] 
\WHILE{$iter \neq 0$}
\STATE Sample a mini batch $X$ from $\mathcal{D}$.
\STATE Pick a group of gates $\alpha^k$ using $k \sim Uniform\{1, ..., K\}$.
\STATE Compute model predictions $\hat{y} = \phi(X|\alpha^k, \Theta)$.
\STATE Optimize network parameters $\Theta$ by Eq. \ref{eq:acc-loss}.
\STATE Optimize gating parameters $\alpha^k$ by Eq. \ref{eq:gating-update}.
\STATE Update $iter$.
\ENDWHILE
\STATE Generate output according to the selected aggregation method.
\STATE \textbf{return} output
\end{algorithmic}
\end{algorithm}

\noindent\textbf{Weight Initialization } We increase the inter-group diversity by randomly initializing the gating parameters according to the following uniform distribution as a means to more evenly spread the gating parameters in the search space and to lower the chance to get trapped in any unexpected local minima:
\small
\begin{equation}
\label{eq:weight-initial}
    \alpha^K_i \sim Uniform(- c \cdot (1-p), c \cdot p).
\end{equation}
\normalsize
Here, $p$ controls the percentage of open gates at the beginning (default value: 0.8), and $c$ determines the magnitude of gate weights (default value: 0.01). Note that Gating and other related NAS methods usually initialize gating parameters to a constant value to make sure every corresponding feature embedding gets considered and fine-tuned at the beginning. However, this is not a concern in Ensemble Gating because for any feature embedding $e_i$, the probability that $\{\alpha_i^k\}$ are all initialized to negative values is very low.\\
\\
\noindent\textbf{Ensemble Aggregation } To combine the decisions of different groups, we propose three candidate result aggregation methods:
\begin{itemize}
    \item Majority Voting (Voting): we sum up the binarized gating decision from all the groups and take the $\mathcal{N}$ features that receive the most votes as the final output.
    \item Averaging (Avg): we compute the average of gating parameters $\{\overline{\alpha}_i=\frac{1}{K} \sum^{k=1}_K \alpha^k_i\}$  and the output is given by the binarized $\{\overline{\alpha}_i\}$.
    \item Minimal Retraining Loss (Min): we retrain $K$ CTR prediction models from scratch with a few training samples (one epoch of training data in our case). Input features are selected according to the results of each group. The final decision is given by the output of the group that yields minimal (average) retraining loss. Note that this aggregation method is more costly than the other two due to the additional model training. 
\end{itemize}

Algorithm \ref{alg:algorithm-ensemble} presents the overall training process of Ensemble Gating. As a summary, we would like to highlight some advantages of Ensemble Gating:
\begin{itemize}
    \item It provides meaningful uncertainty-driven exploration in order to search for the best subset of features.
    \item Considering the number of input features is much smaller than the number of parameters in a deep CTR prediction model, creating multiple gating layers doesn't significantly increase memory usage. Hence, Ensemble Gating is a scalable method. Also, its parallelizable nature makes it possible to simultaneously update multiple groups of binary gates, and thereby making it well suited for any distributed learning framework. 
    \item Random weight initialization and random group selection (randomly selecting one group of gates at each update step) work as additional intrinsic randomization to lower the chance of getting stuck into any local minima.
    \item Ensemble Gating doesn't make any assumption on the underlying gradient-based feature selection method. Although in this paper we use it to improve the Gating method, it can also apply to other methods with minimum adaptation.
\end{itemize}

%% file: 5-experiment.tex
\section{Experiments}
\label{sec:expr}

\begin{table*}[t]
\fontsize{8}{8}\selectfont
\centering
\begin{tabular}{p{0.05\linewidth} p{0.05\linewidth} p{0.08\linewidth} p{0.07\linewidth} p{0.07\linewidth} p{0.07\linewidth} p{0.07\linewidth} p{0.07\linewidth}}
\toprule
Dataset & Model & All Features & Random & Gating & Ensemble Avg & Ensemble Voting & Ensemble Min \\
\midrule
\multirow{3}{0.1\linewidth}{Criteo} & DCN & 81.37 & 79.857 (0.0052) & 80.725 (0.0023) & 81.014 (0.0001) & 81.006 (0.0003) & \textbf{81.028 (0.0001)}\\\cmidrule{2-8}
& AutoInt & 81.28 & 79.273 (0.0061) & 80.673 (0.0013) & 80.967 (0.0003) & \textbf{80.983 (0.0003)} & 80.938 (0.0005)\\\cmidrule{2-8}
& DeepFM & 80.43 & 78.737 (0.0031) & 80.22 (0.0011) & 80.336 (0.0006) & 80.29 (0.0005) & \textbf{80.336 (0.0005)} \\
\midrule
\multirow{3}{0.1\linewidth}{Avazu} & DCN & 78.72 & 73.527 (0.0195) & 77.612 (0.0052) & 78.178 (0.001) & 78.184 (0.001) & \textbf{78.184 (0.0007)}\\\cmidrule{2-8}
& AutoInt & 78.81 & 72.543 (0.0039) & 77.882 (0.0034) & \textbf{78.17 (0.0007)} & 78.17 (0.0007) & 78.162 (0.0009)\\\cmidrule{2-8}
& DeepFM & 78.0 & 74.307 (0.0126) & 75.59 (0.0286) & 77.592 (0.001) & 77.404 (0.0048)) & \textbf{77.618 (0.0008)}\\
\bottomrule
\end{tabular}
\caption{Average AUC (\%) and standard deviation on testing set. Ensemble Gating outperforms Gating on the two benchmark dataset with three different deep CTR prediction models.}
\label{table:all-results}
\normalsize
\end{table*}

In this section, we present the experiments that evaluate the effectiveness of Ensemble Gating. Firstly, we intend to investigate if our ensemble method can consistently improve Gating. To this end, we compare Ensemble Gating against Gating on two public benchmark datasets (Avazu\footnote{https://www.kaggle.com/c/avazu-ctr-prediction} and Criteo\footnote{https://www.kaggle.com/c/criteo-display-ad-challenge}) with three different underlying deep CTR prediction models, namely DeepFM \cite{guo2017deepfm}, AutoInt \cite{song2019autoint}, and DCN \cite{wang2017deep}. Note that depending on the ensemble aggregation method used, there can three different implementations of Ensemble Gating, which are referred to as Ensemble Avg, Ensemble Voting, and Ensemble Min, respectively. Then, to get more insights into Ensemble Gating, we perform a set of complementary experiments, such as varying the numbers of gating layers (ensemble size), and initializing gating parameters in different ways. \\
\\
\noindent \textbf{Datasets and Tasks} 
To comply with the policies of most companies, we refrain from presenting the empirical results obtained on any industrial large-scale dataset. Instead, we chose to use two largest open CTR prediction datasets, namely Avazu and Criteo. In Avazu, there are 22 feature fields and the task is to select 6 features that have the highest explanatory power. In Criteo, there are totally 39 feature fields and the task is to select a subset of 19 features. 

We randomly split the dataset into 80\% and 20\% for training and testing. The testing dataset is further randomly split in half for validation and testing in our complementary experiments where network parameter update and gating parameter update are performed on different subsets of data. The evaluation metric we adopt is the AUC (Area Under the ROC Curve) achieved on the testing set by the model that is retrained with the selected features.\\
\\
\noindent \textbf{Training Setup}
Both Ensemble Gating and the baselines follow the same training process. Firstly, we pretrain a CTR prediction model with all input features for 10 epochs. We apply the Adam optimizer \cite{kingma2015adam} with a learning rate of 0.001 for network parameters $\Theta$. The mini-batch size is set to 2048 and the weight of weight decay regularizer is set to $1e^{-6}$. The weights of pretrained model are saved for later steps. 

After obtaining the pretrained model, the gating layer(s) will be inserted between the embedding layers and the interaction layers. Then we iteratively updated the network parameters $\Theta$ and the gating parameters $\alpha$ for 8 epochs. We use SGD with a learning rate of 0.001 to optimize the gating parameters. 

Following the Lottery Ticket Hypothesis \cite{frankle2018lottery}, our last step is to retrain the model from scratch with selected features on the training set (for 10 epochs) and test it on the testing set. The final gating decision of Ensemble Gating is calculated according to the particular aggregation method. In our preliminary experiments, we observed that in both Gating and Ensemble Gating, the number of open gates at the end of training is occasionally one or two smaller than the target value $\mathcal{N}$. The most likely cause of this phenomenon is the choice of the hyperparameter $\beta_s$ (weight of sparse regularizer).  Considering that finding the ``optimal" value of $\beta_s$ can be very expensive in practice, we adopt a simple workaround, which adds unselected features with the largest gate weights (average gate weights in ensemble method) to the set of selected features. This process resembles adjusting the gating threshold that is set to zero in the binarized function (Eq. \ref{eq:ste-decision}). Our preliminary results suggest that this post-processing doesn't have any observable negative impact on the final outcome. 

\subsection{Experiment Results}
We ran each feature selection method 5 times and recorded the AUC score achieved on the testing set. Tab. \ref{table:all-results} and Fig. \ref{fig:all-results} in Appendix A present the results of Ensemble Gating and the baselines. For reference, we also include the results of the random baseline (randomly selecting $\mathcal{N}$ features) and the full-features baseline (using all the features in retraining). The results show that both Gating and Ensemble Gating outperform the random baseline by a large margin, and Ensemble Gating is able to further improve Gating by selecting a better subset of features that results in a significantly higher average test AUC. Moreover, Ensemble Gating also has a noticeably smaller standard deviation of AUC scores, which suggests that Ensemble Gating's uncertain-driven exploration enables it to more consistently find a set of informative features regardless of all the randomization (e.g. dataset shuffling) and its initial position in the search space. 

\begin{figure}
\begin{center}
\centerline{\includegraphics[width=0.7\columnwidth]{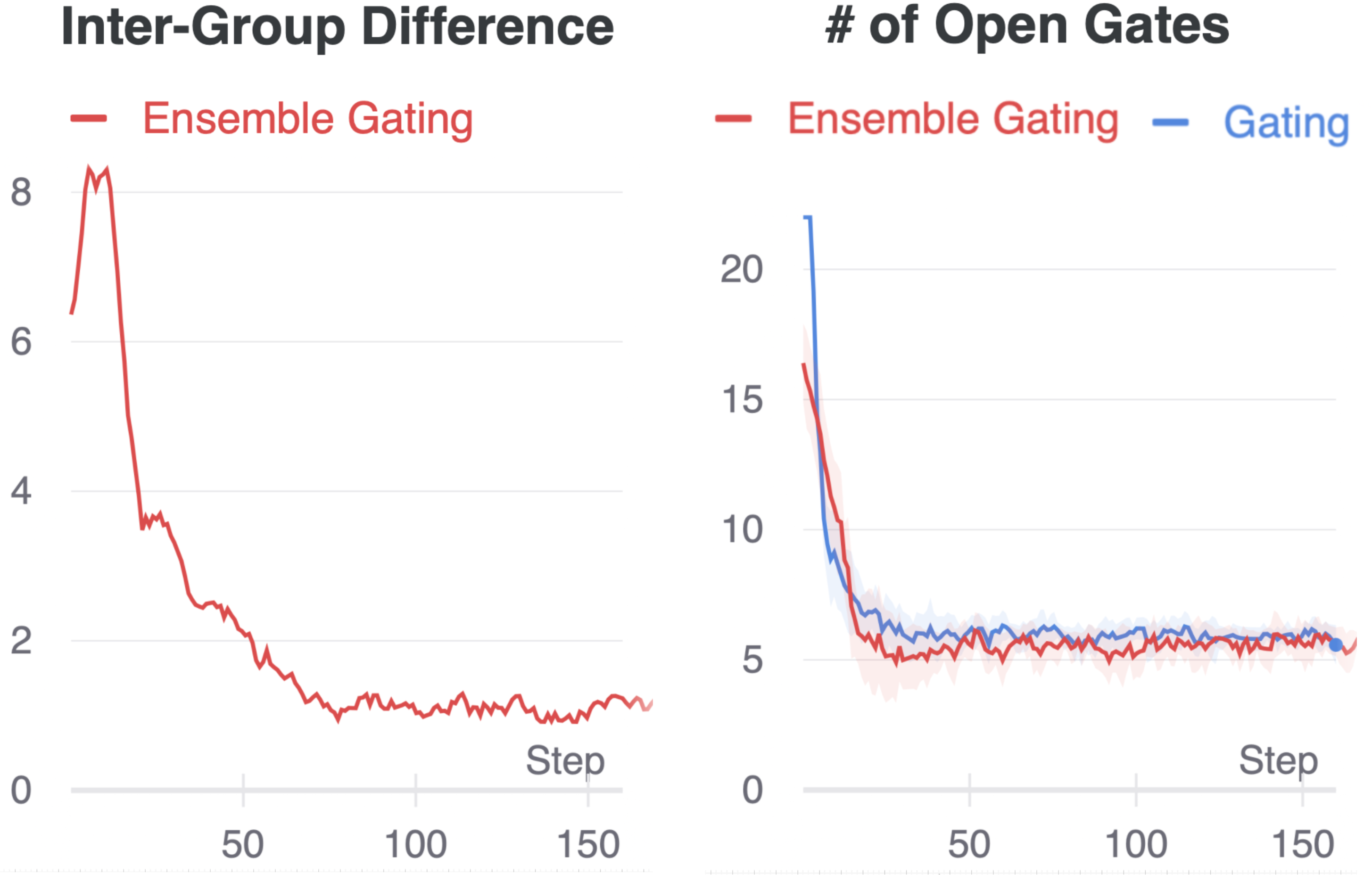}}
\caption{The inter-group difference (left) and the number of open gates (right) during searching process on Avazu. The logging frequency was set to 500, so every logging step in above figures corresponds to 500 parameter update steps. The results verify that both Ensemble Gating and Gating converge within 2 epochs (around 100 logging steps).}
\label{fig:group-diff}
\end{center}
\vspace{-0.5cm}
\end{figure}

One might argue that the mean and standard deviation do not tell the whole story, because when there are adequate computational resources, people can run the searching algorithm for multiple times and pick the best one as the final output. To address this concern, we rank the results of Gating and Ensemble Avg according to their AUC scores, and count how many of the top three results were obtained with the ensemble method. We refer to this value as Top-3 score. On Criteo, our Top-3 scores are 3 (DCN), 3 (AutoInt), and 2 (DeepFM); on Avazu, the scores are 3 (DCN), 3 (AutoInt), and 3 (DeepFM). This confirms that Ensemble Gating exhibits better exploratory capacity and manages to constantly find a better subset of features. \\ 
\\
\noindent \textbf{Convergence Discussion } As pointed out by many previous works \cite{Xiao2019AutoPruneAN, gordon2018morphnet}, there is no strong guarantee of convergence in Gating, but empirical results suggest that Gating usually converges to a fairly well state within a few epochs in practice. Hence, in this section, we intend to examine whether this fast ``convergence" property holds in our ensemble method. To this end, we keep track of the average number of open gates and inter-group differences in the searching/training process. The inter-group difference is measured by counting the number of features on which two groups disagree. Here, we take the results obtained on Avazu with AutoInt as example (Fig. \ref{fig:group-diff}). The plots show that the number of open gates is quickly optimized to the target value and the inter-group difference drops to a reasonably small value within two epochs. Note that theoretically the inter-group difference is not necessarily zero at convergence because of the non-convex nature of the feature selection problem and the fact that there might exist multiple optimal solutions. \\

\subsection{Additional Experiments}
In this section, we will conduct a set of experiments with the aim of getting more insights into Ensemble Gating. We assume the Average aggregation method is used in all the complementary experiments. \\

\noindent \textbf{What is a good value for the size of ensemble? } The answer to this question can be found in Fig. \ref{fig:all-results} in Appendix A, in which we vary the number of binary gate groups. It is clear that the performance of 5 groups and 10 groups is better than that of 2 groups, and 2 groups outperforms Gating in only a few cases. Hence, we can conclude that an ensemble of size 5 is sufficient to capture the benefits of the ensemble method. \\
\\
\noindent \textbf{Is our uncertainty-driven exploration a better strategy compared to other methods? } To answer this question, we compare Ensemble Gating to an approach that uses another popular binarize function Gumbel-Softmax \cite{jang2016categorical}. Since each binary gate only has two states (open or closed), Gumbel-Softmax function in this case is essentially a Gumbel-Sigmoid function as in \cite{tsai2018learning}:
\small
\begin{equation}
\label{eq:gumbel}
    binarize\texttt{\_}Gumbel(\alpha_i) = sigmoid(\frac{\alpha_i + g_k - g_l}{\gamma}),
\end{equation}
\normalsize
where $g=-log(-log(u))$ is a Gumbel noise, $u$ is a uniform random variable in the range of $(0, 1)$, and $\gamma$ is a small enough temperature parameter. Gumbel-Sigmoid function becomes closer to a true binarize function when $\gamma$ approaches zero. Similar to \cite{xie2018snas}, we gradually decrease the value of $\gamma$ from $1e^{-3}$ to $1e^{-4}$ throughout the searching/training process. We also experimented with other temperature annealing configurations, but results remained similar. 
Gumbel-Sigmoid function can potentially improve gating exploration due to the existence of the Gumbel noise. It resembles $\epsilon$-greedy exploration in Reinforcement Learning in a sense that both of them add random exploration incentive without considering state context. On the contrary, Ensemble Gating's exploration strategy is based on predictive uncertainty, which is highly state dependent. Tab. \ref{table:gumbel-results} compares the performance of Ensemble Gating and Gumbel-Sigmoid, and it shows that uncertainty-driven exploration works better than simply adding random noise into the system. \\

\begin{table}[t]
\fontsize{7}{7}\selectfont
\centering
\begin{tabular}{p{0.1\linewidth}p{0.12\linewidth} p{0.12\linewidth}p{0.12\linewidth}p{0.12\linewidth}}
\toprule
Dataset & Model & Gumbel-Sigmoid & Gating & Ensemble Avg  \\
\midrule
\multirow{3}{0.1\linewidth}{Criteo} & DCN & 80.427 (0.0026) & 80.725 (0.0023) & \textbf{81.014 (0.0001)} \\\cmidrule{2-5}
& AutoInt & 80.472 (0.0011) & 80.673 (0.0013) & \textbf{80.967 (0.0003)} \\\cmidrule{2-5}
& DeepFM & 79.813 (0.0017) & 80.22 (0.0011) & \textbf{80.336 (0.0006)} \\
\midrule
\multirow{3}{0.1\linewidth}{Avazu} & DCN & 77.254 (0.0049) & 77.612 (0.0052) & \textbf{78.178 (0.001)} \\\cmidrule{2-5}
& AutoInt & 77.325 (0.0075) & 77.882 (0.0034) & \textbf{78.17 (0.0007) }\\\cmidrule{2-5}
& DeepFM & 76.636 (0.0065) & 75.59 (0.0286) & \textbf{77.592 (0.001)} \\
\bottomrule
\end{tabular}
\caption{Comparison between Ensemble Gating and Gumbel-Sigmoid. The results are visualized in Fig. \ref{fig:appendix-gumbel-criteo} and Fig. \ref{fig:appendix-gumbel-avazu} in Appendix A.}
\label{table:gumbel-results}
\vspace{-0.3cm}
\normalsize
\end{table}

\noindent \textbf{Is random weight initialization employed by Ensemble Gating a key factor that distinguishes Ensemble Gating from Gating? } One difference between Gating and Ensemble Gating is the way they initialize the gate weights. Here we perform additional experiments to examine the effect of different initialization methods on the final outcome. We refer to Ensemble Gating with random initialization as \textit{Ensemble Random}, and our ensemble method with constant initialization as \textit{Ensemble Constant}. We follow the same naming convention for Gating. The results are presented in Tab. \ref{table:init-results}. We can observe that the performance of Ensemble Random and Ensemble Constant are very close while Ensemble Random still maintains a slight edge over other approaches. Not surprisingly, random weight initialization degrades the performance of Gating. This can be explained by the gating overfitting problem too: the network parameters are fine-tuned to prefer features with gates that are initialized to be open. On the contrary, as mentioned in the previous section, having a relatively small number of gates closed initially is not a concern in ensemble method. Hence, Ensemble Gating can benefit from the inter-group diversity that arises from random initialization.   \\

\begin{table}[t]
\centering
\fontsize{7}{7}\selectfont
\begin{tabular}{p{0.07\linewidth}p{0.11\linewidth}p{0.11\linewidth}p{0.12\linewidth}p{0.11\linewidth}p{0.12\linewidth}}
\toprule
  & Model & Gating Constant & Gating Random & Ensemble Random & Ensemble Constant  \\
\midrule
\multirow{3}{0.1\linewidth}{Criteo} & DCN & 80.725 (0.0023) & 80.516 (0.0042) & \textbf{81.014 (0.0001)} & 80.96 (0.0005) \\\cmidrule{2-6}
& AutoInt & 80.673 (0.0013) & 80.74 (0.0018) & \textbf{80.967 (0.0003)} & 80.93 (0.0004) \\\cmidrule{2-6}
& DeepFM & 80.22 (0.0011) & 80.133 (0.0013) & \textbf{80.336 (0.0006)} & 80.273 (0.0007) \\
\midrule
\multirow{3}{0.1\linewidth}{Avazu} & DCN & 77.612 (0.0052) & 77.504 (0.0033) & \textbf{78.178 (0.001)} & 78.125 (0.001) \\\cmidrule{2-6}
& AutoInt & 77.882 (0.0034) & 77.122 (0.0125) & 78.17 (0.0007) & \textbf{78.195 (0.0001)}\\\cmidrule{2-6}
& DeepFM & 75.59 (0.0286) & 74.39 (0.0199) & \textbf{77.592 (0.001)} & 77.677 (0.0005) \\
\bottomrule
\end{tabular}
\caption{Comparison of different weight initialization methods. The results are visualized in Fig. \ref{fig:appendix-init-criteo} and Fig. \ref{fig:appendix-init-avazu} in Appendix A.}
\label{table:init-results}
\vspace{-0.3cm}
\normalsize
\end{table}

\noindent \textbf{How important is it to update the binary gates on a separate validation set? } Some related works suggest training gating parameters on a separate validation dataset in order to enhance the generalization of the gating result \cite{zhao2021autodim, Xiao2019AutoPruneAN, cai2018proxylessnas}. However, in our experiments, the performance of Gating doesn't get improved when $\mathcal{D}_g$ is different from $\mathcal{D}$ (see Appendix A for all the results). One possible reason for this is the validation set in Avazu and Criteo might be not large enough to capture the distribution of all the data. Nevertheless, Ensemble Gating is able to outperform Gating under both settings (i.e. $\mathcal{D}=\mathcal{D}_g$ and $\mathcal{D}\neq\mathcal{D}_g$). \\

\noindent \textbf{Can we freeze network parameters to prevent the gating overfitting issue?  } Most gradient-based NAS and network pruning methods follow the multi-step training/searching framework, which iteratively updates the sparsity structure and fine-tune the original model weights. We show in previous section that the major drawback of this multi-step training framework is the model weights can end up ``overfitting" to certain gating state. Hence, one natural question is whether we can bypass the model fine-tuning to avoid the gating overfitting issue. Our experimental results confirm that fine-tuning the network parameters is an indispensable step to find a good subset of features (see Appendix A for all the results). Hence, the right way to improve differentiable neural feature selection methods should be looking for better exploration strategy rather than bypassing the fine-tuning steps.

%% file: 6-conclusion.tex
\section{Conclusion \& Future Work}
In this work, we presented a novel ensemble method that offers a more efficient exploration strategy to find the optimal subset of features. We show that our proposed method, Ensemble Gating, consistently improves Gating method on two public datasets with three different underlying CTR prediction models. The proposed method is computationally efficient and can be easily applied to a variety of CTR prediction models. 

Future work may include studying whether our ensemble method can benefit related research problems, such as differentiable network pruning and neural input search. Insufficient exploration can be detrimental in any search problem, and gating overfitting may exist as long as similar gating formulations are employed. Moreover, we note that we have used our ensemble method to improve Gating method implemented with step function and STE. Future work can experiment with other binarize functions.

%% file: 7-appendix.tex
\clearpage
\appendix
\section{Appendix}

\subsection{A. Results of All Experiments}
In this section, we present all the experimental results, some of which were not included in the paper due to space constraint.

\subsubsection{Testing AUC scores of Ensemble Gating and Gating}
\begin{figure*}[th!]
\centering
\subfloat[Avazu, DCN]{\includegraphics[width=0.166\linewidth]{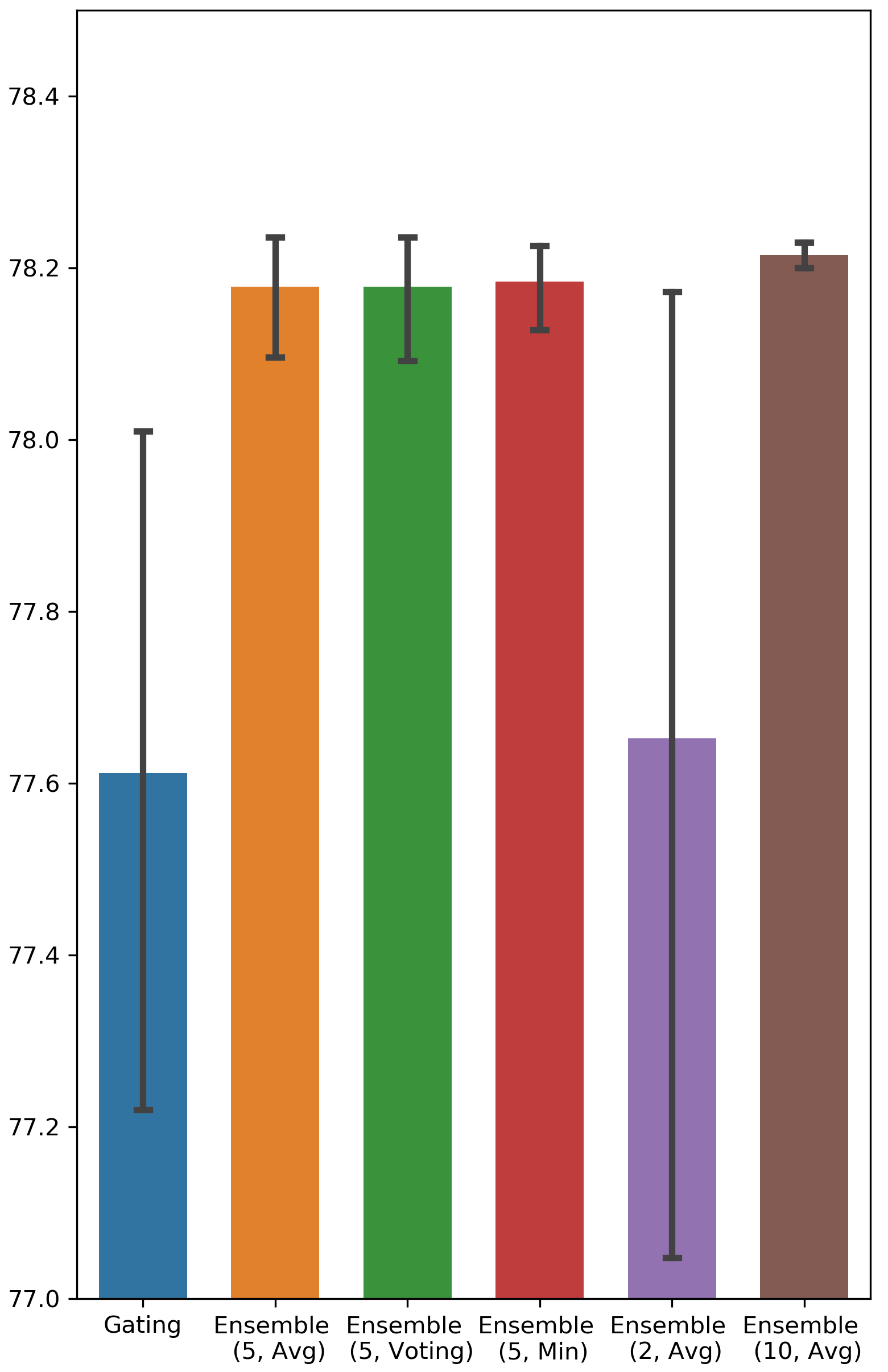}}
\subfloat[Avazu, AutoInt]{\includegraphics[width=0.166\linewidth]{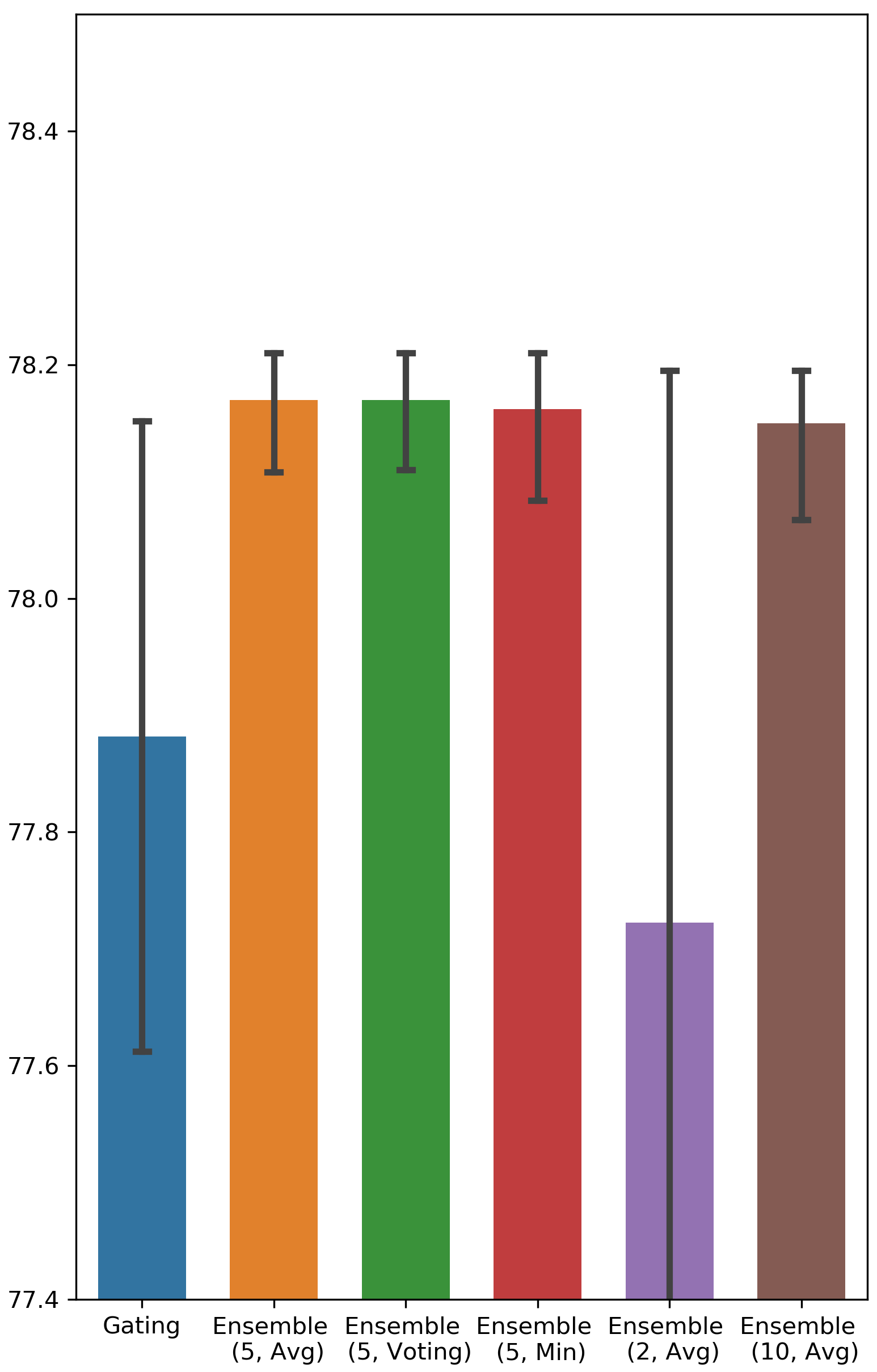}}
\subfloat[Avazu, DeepFM]{\includegraphics[width=0.166\linewidth]{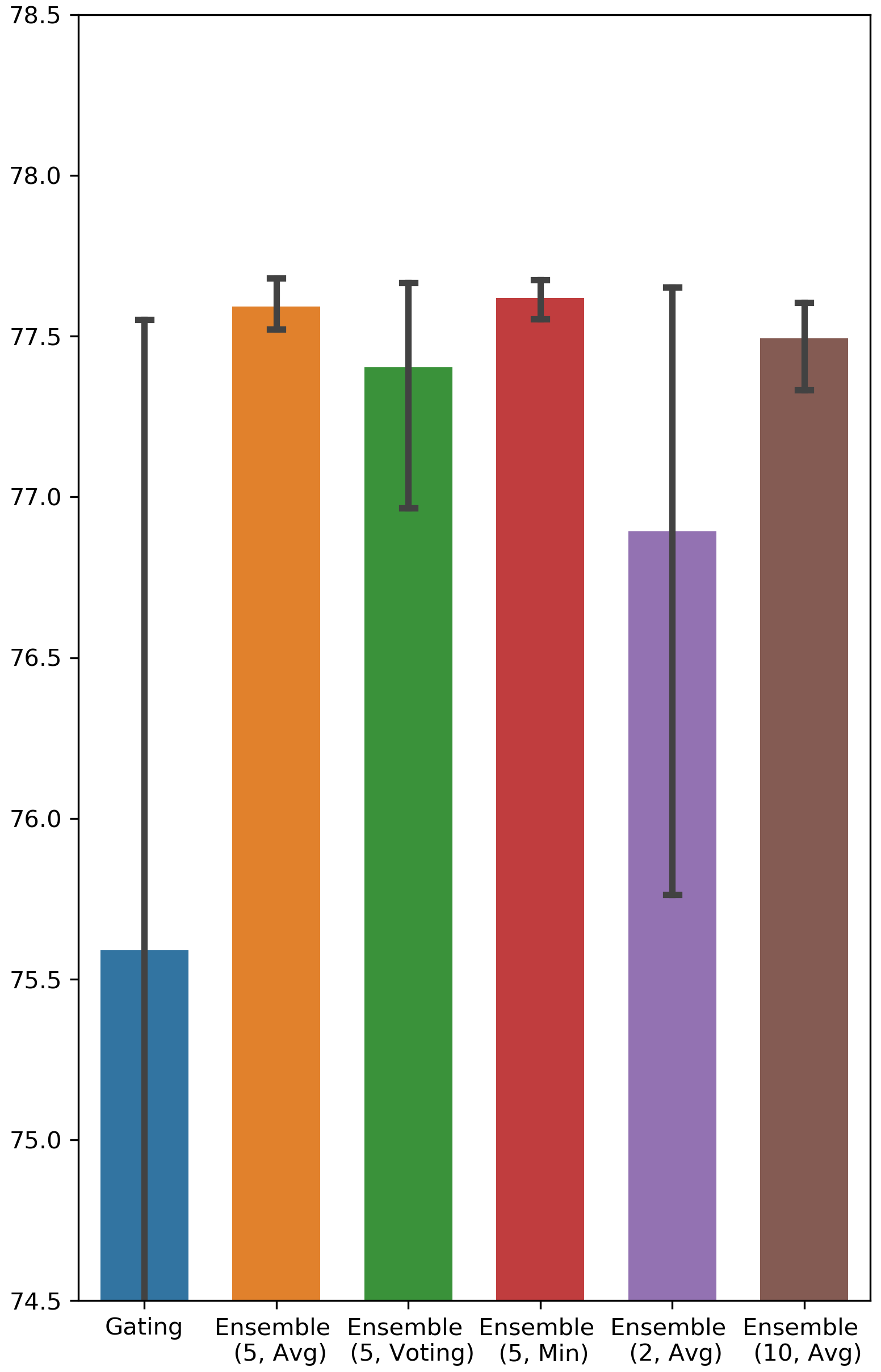}}
\subfloat[Criteo, DCN]{\includegraphics[width=0.166\linewidth]{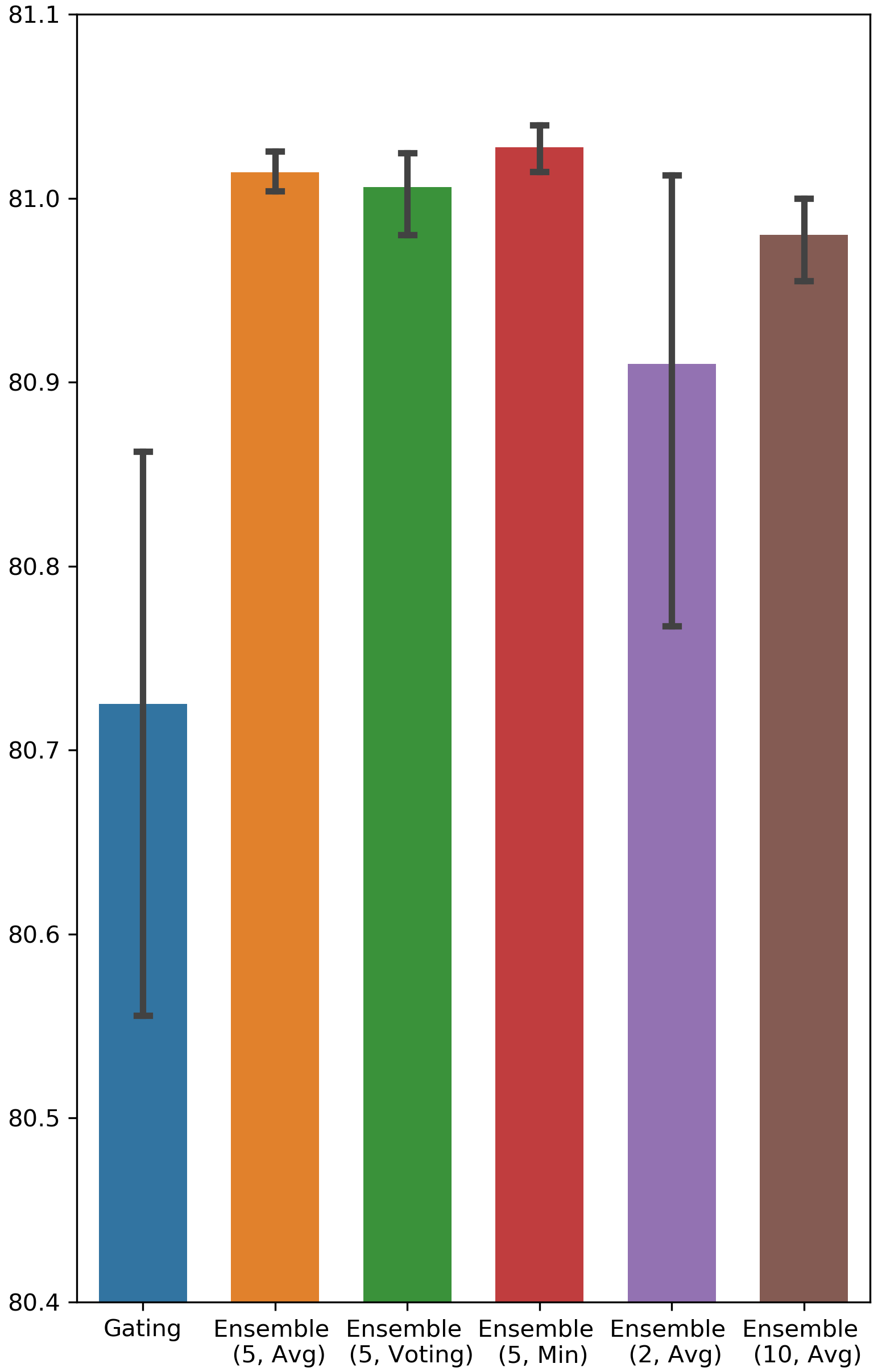}}
\subfloat[Criteo, AutoInt]{\includegraphics[width=0.166\linewidth]{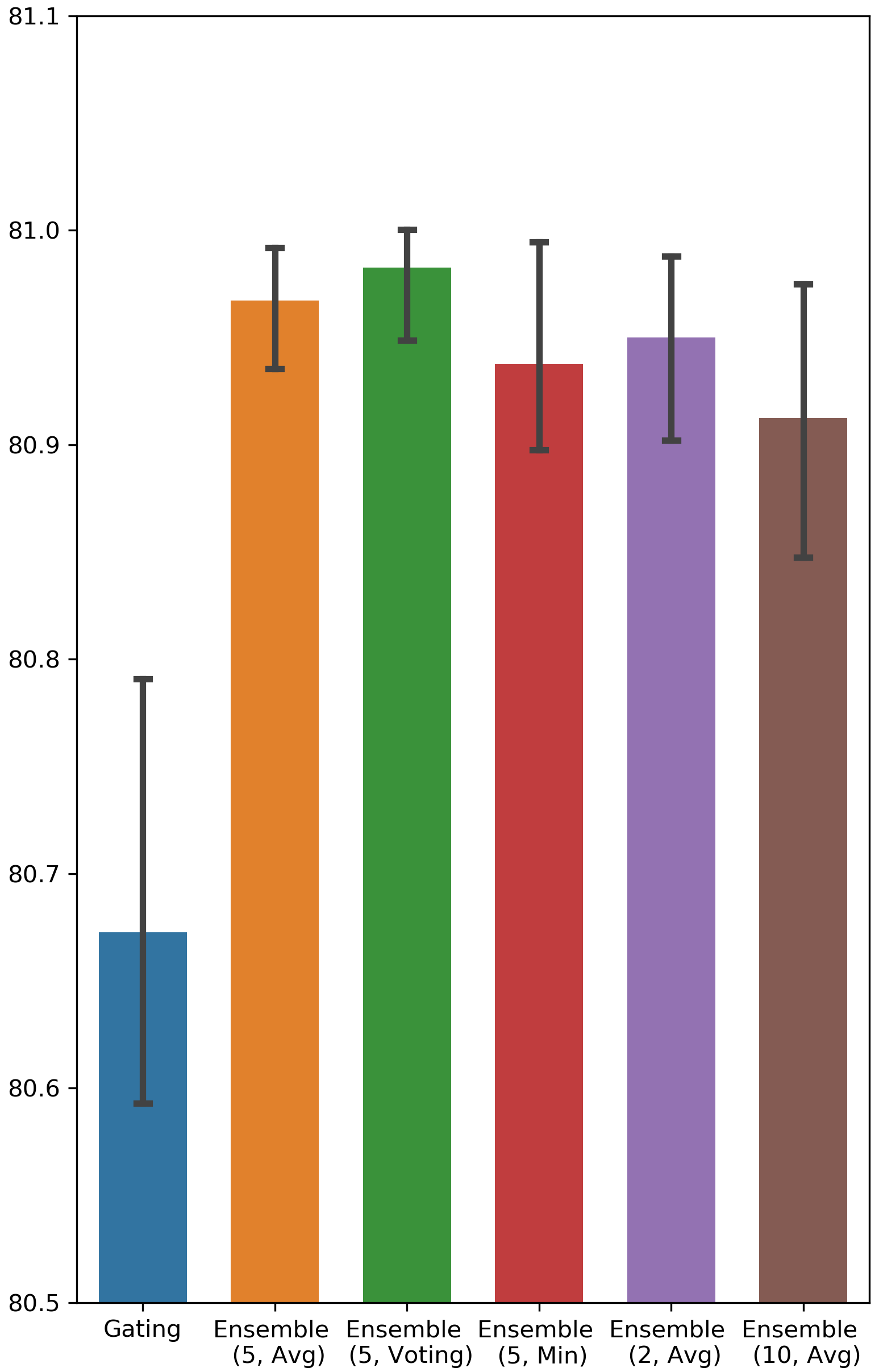}}
\subfloat[Criteo, DeepFM]{\includegraphics[width=0.166\linewidth]{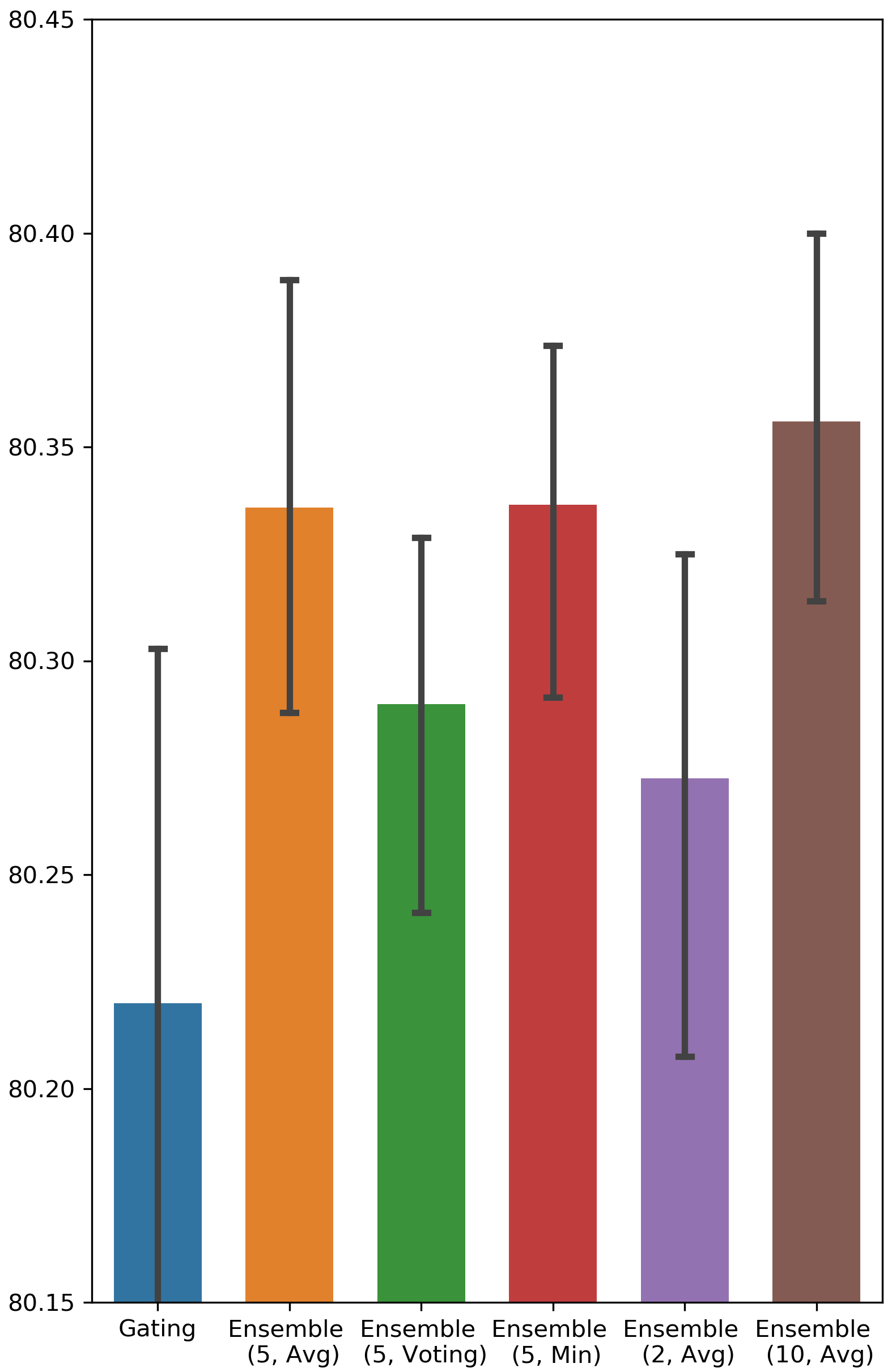}}
\caption{Comparison of different gradient-based feature selection techniques. For each sub-plot, from left to right, each bar corresponds to Gating (blue), Ensemble Gating using Avg aggregation method (orange), Ensemble Gating using Voting aggregation method (green), Ensemble Gating using Min aggregation method (red), Ensemble Gating using two groups of gates (purple), and Ensemble Gating using ten groups of gates (brown), respectively.}
\label{fig:all-results}
\end{figure*}

The results are visualized in Fig. \ref{fig:all-results}.

\subsubsection{Comparison between Ensemble Gating and Gumbel-Sigmoid } 
\begin{figure}[!ht]
\centering
\subfloat[DCN]{\includegraphics[width=0.3\linewidth]{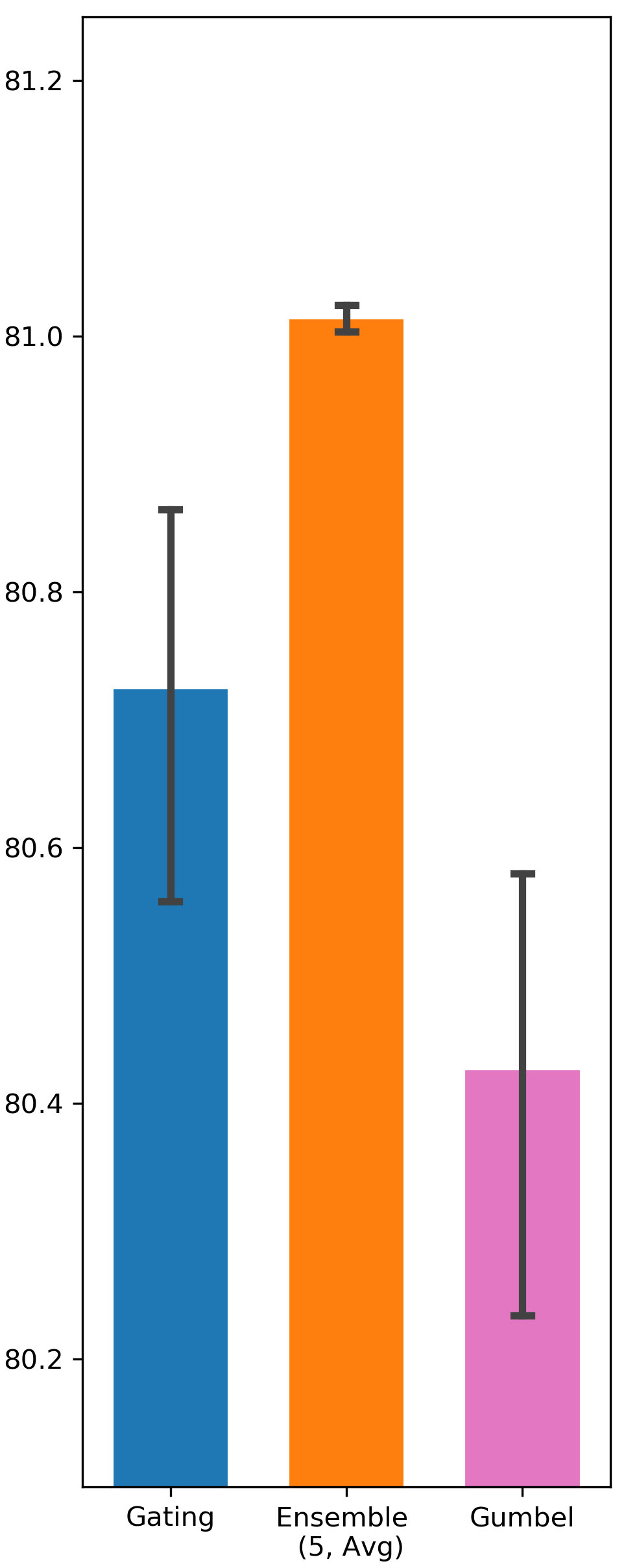}}
\subfloat[AutoInt]{\includegraphics[width=0.3\linewidth]{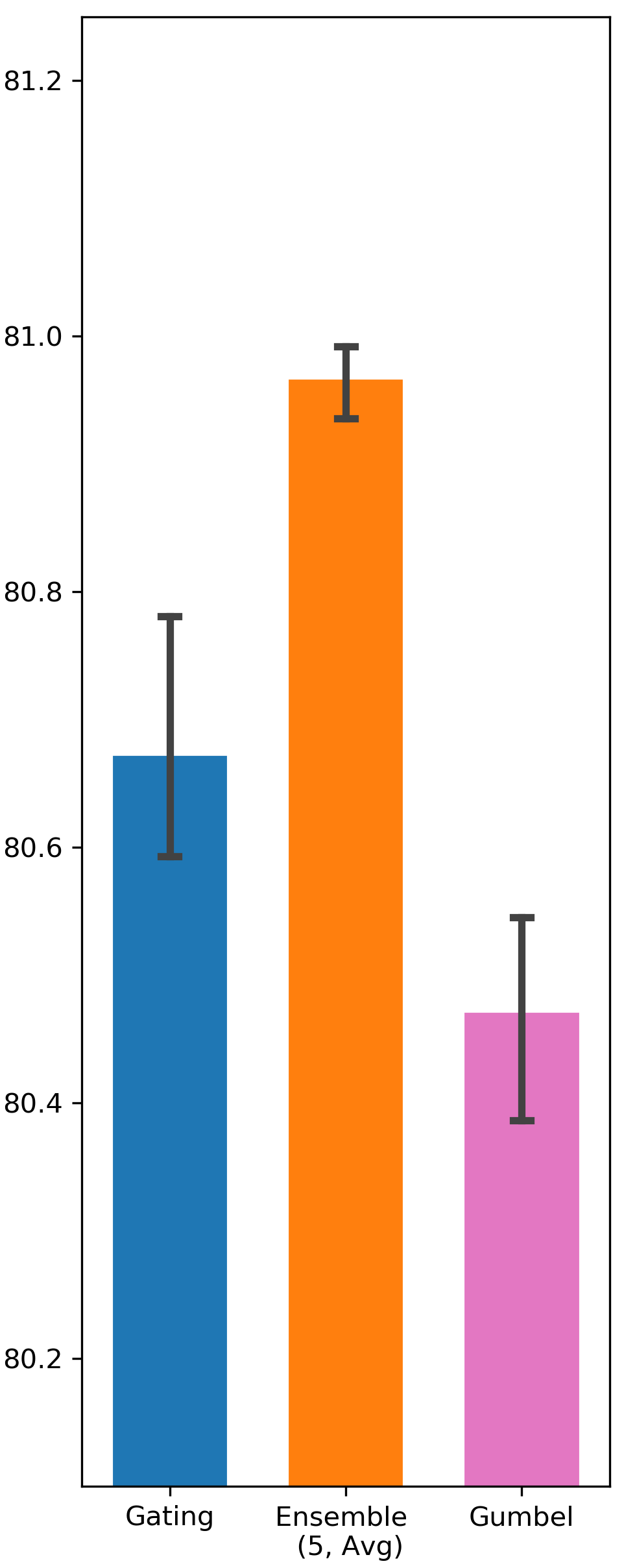}}
\subfloat[DeepFM]{\includegraphics[width=0.3\linewidth]{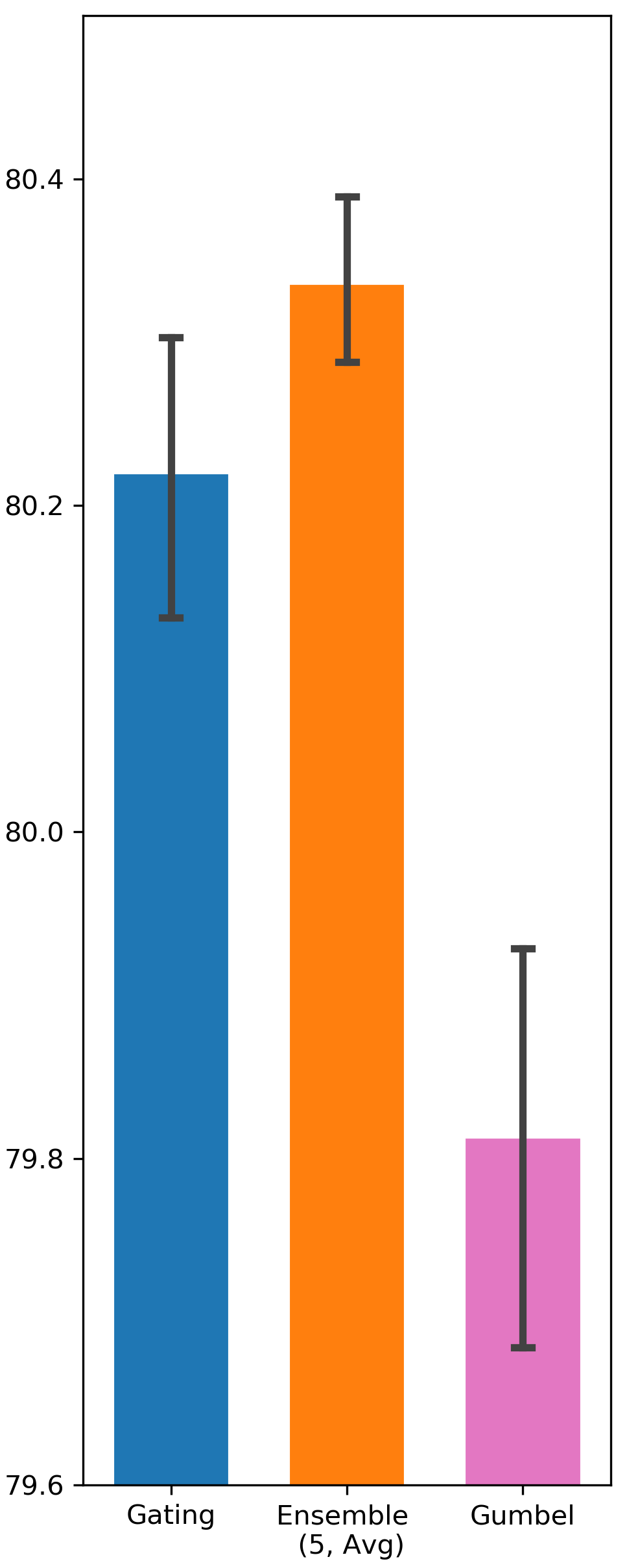}}
\caption{Comparison between Ensemble Gating and Gumbel-Sigmoid on Criteo. The blue bar represents Gating, the orange bar represents Ensemble Gating, and the pink bar represents Gumbel-Sigmoid.}
\label{fig:appendix-gumbel-criteo}
\end{figure}

\begin{figure}[!ht]
\centering
\subfloat[DCN]{\includegraphics[width=0.3\linewidth]{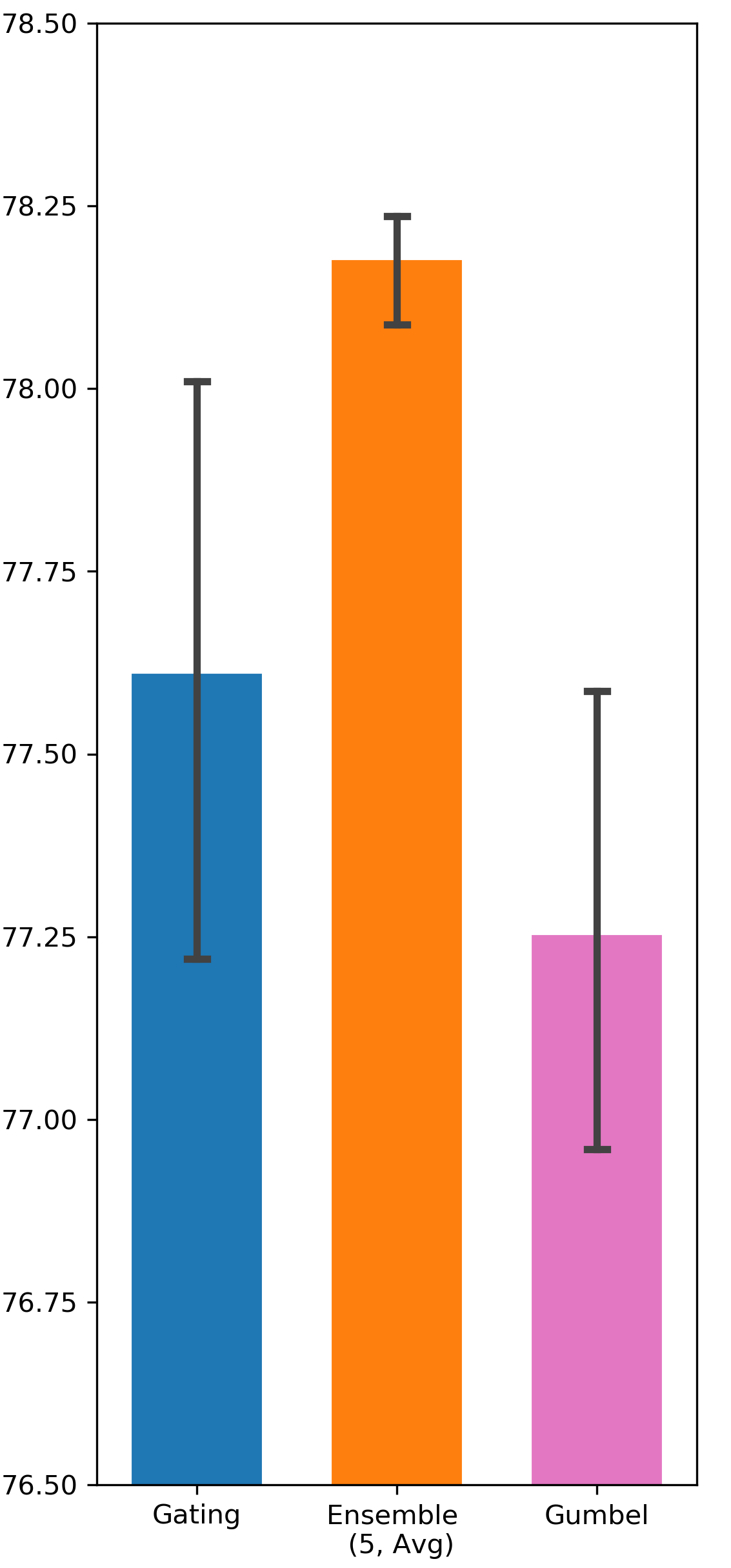}}
\subfloat[AutoInt]{\includegraphics[width=0.3\linewidth]{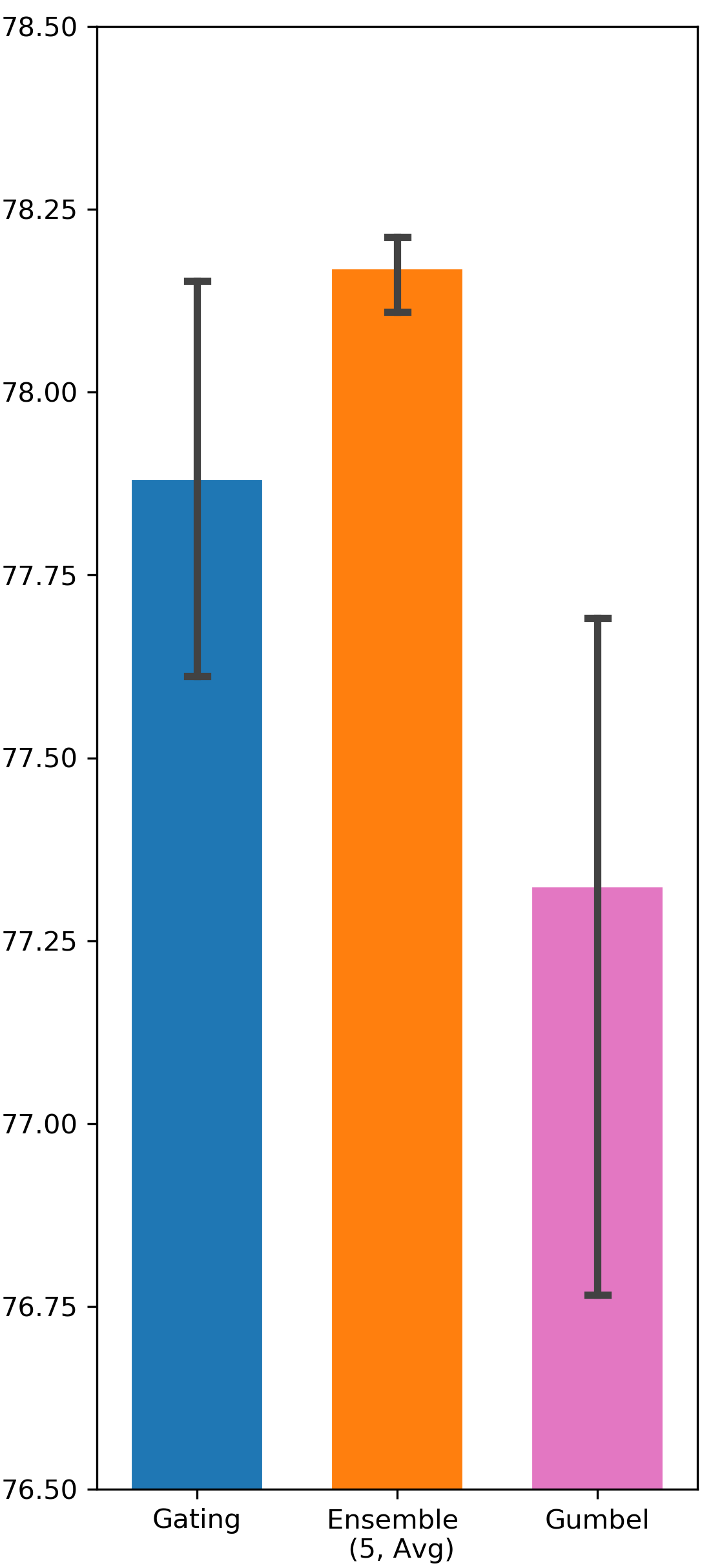}}
\subfloat[DeepFM]{\includegraphics[width=0.3\linewidth]{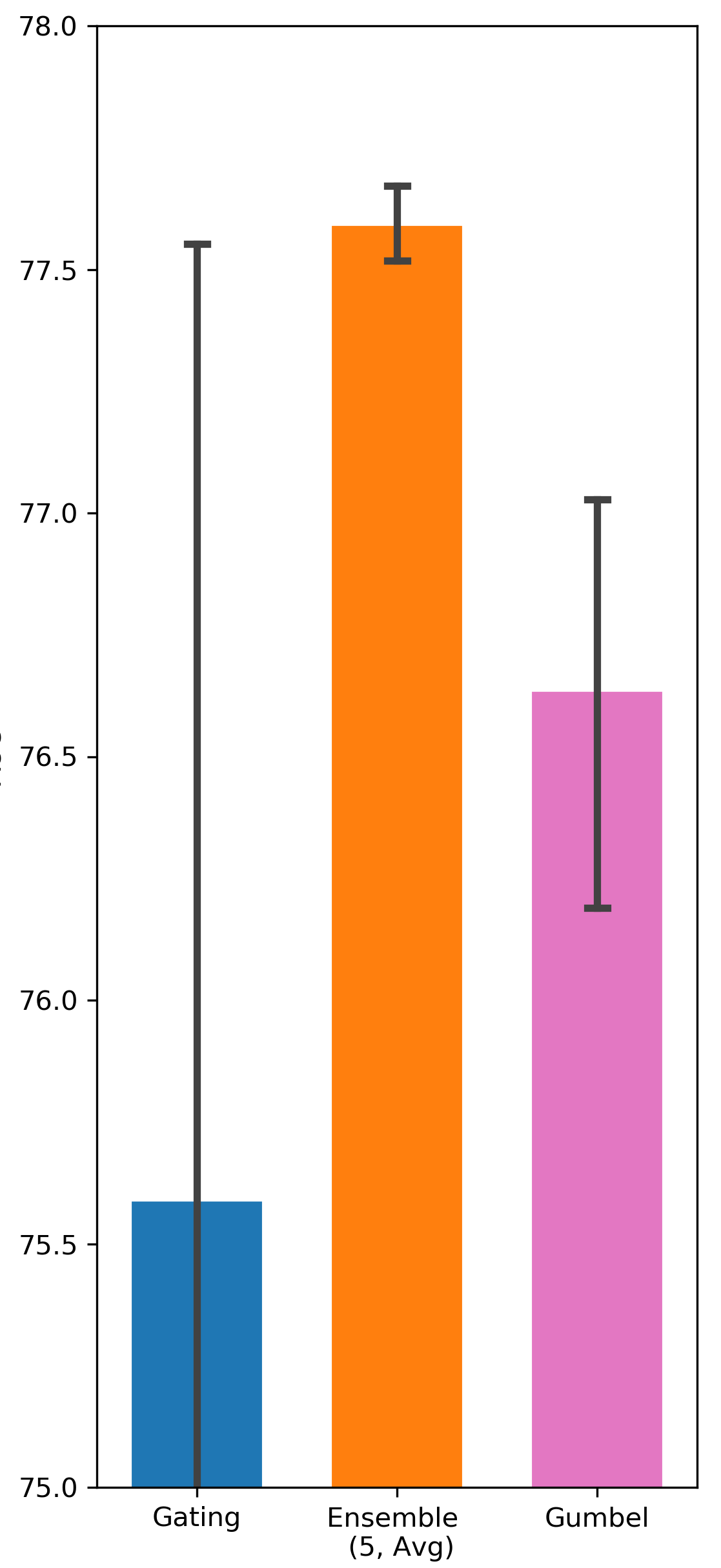}}
\caption{Comparison between Ensemble Gating and Gumbel-Sigmoid on Avazu. The blue bar represents Gating, the orange bar represents Ensemble Gating, and the pink bar represents Gumbel-Sigmoid.}
\label{fig:appendix-gumbel-avazu}
\end{figure}

We intend to verify whether the uncertainty-driven exploration can be more efficient than other exploration strategy. The results are visualized in Fig. \ref{fig:appendix-gumbel-criteo} and Fig. \ref{fig:appendix-gumbel-avazu}.

\subsubsection{Comparison between Different Weights Initialization Methods } 
\begin{figure}
\centering
\subfloat[DCN]{\includegraphics[width=0.3\linewidth]{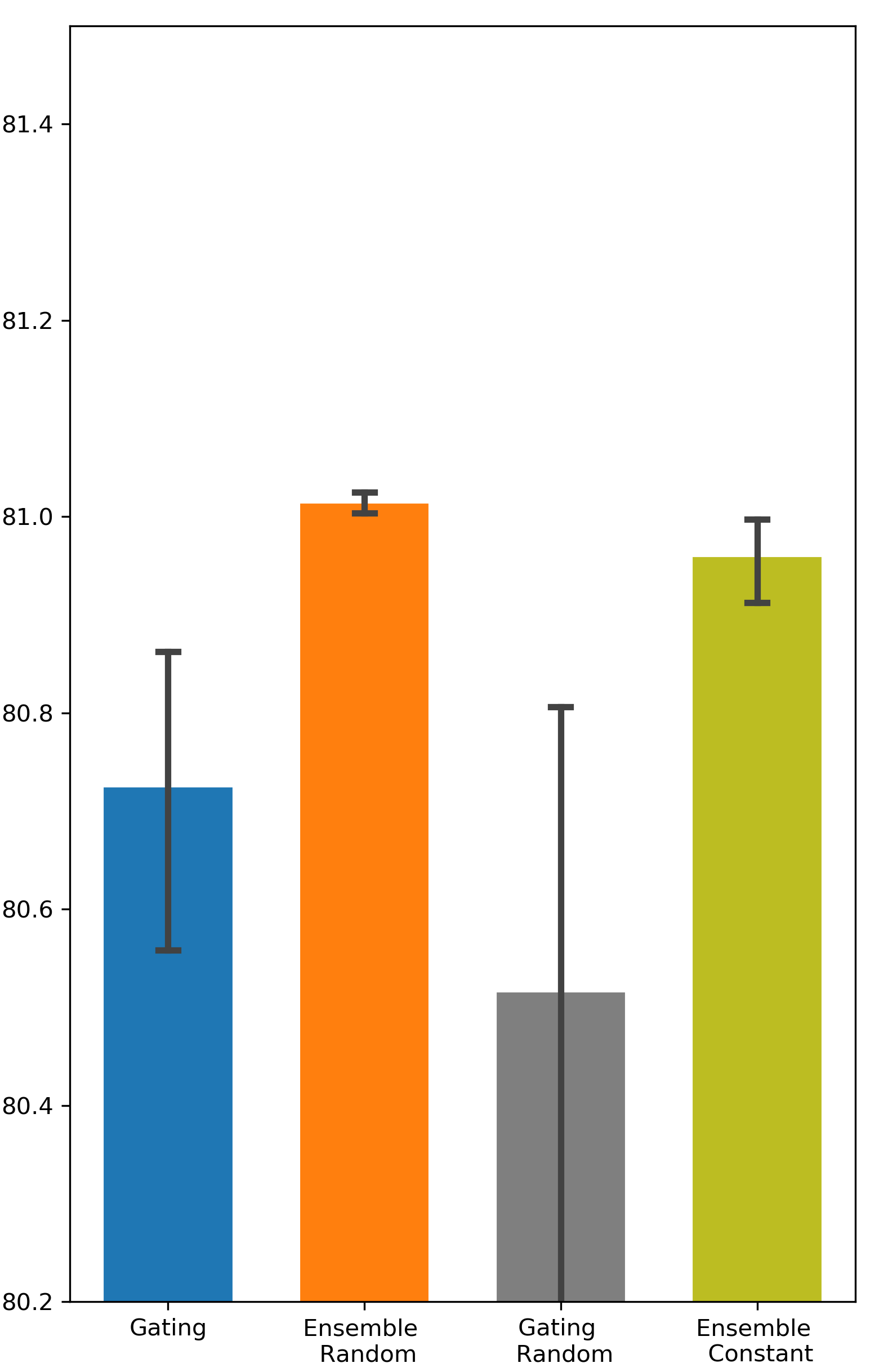}}
\subfloat[AutoInt]{\includegraphics[width=0.3\linewidth]{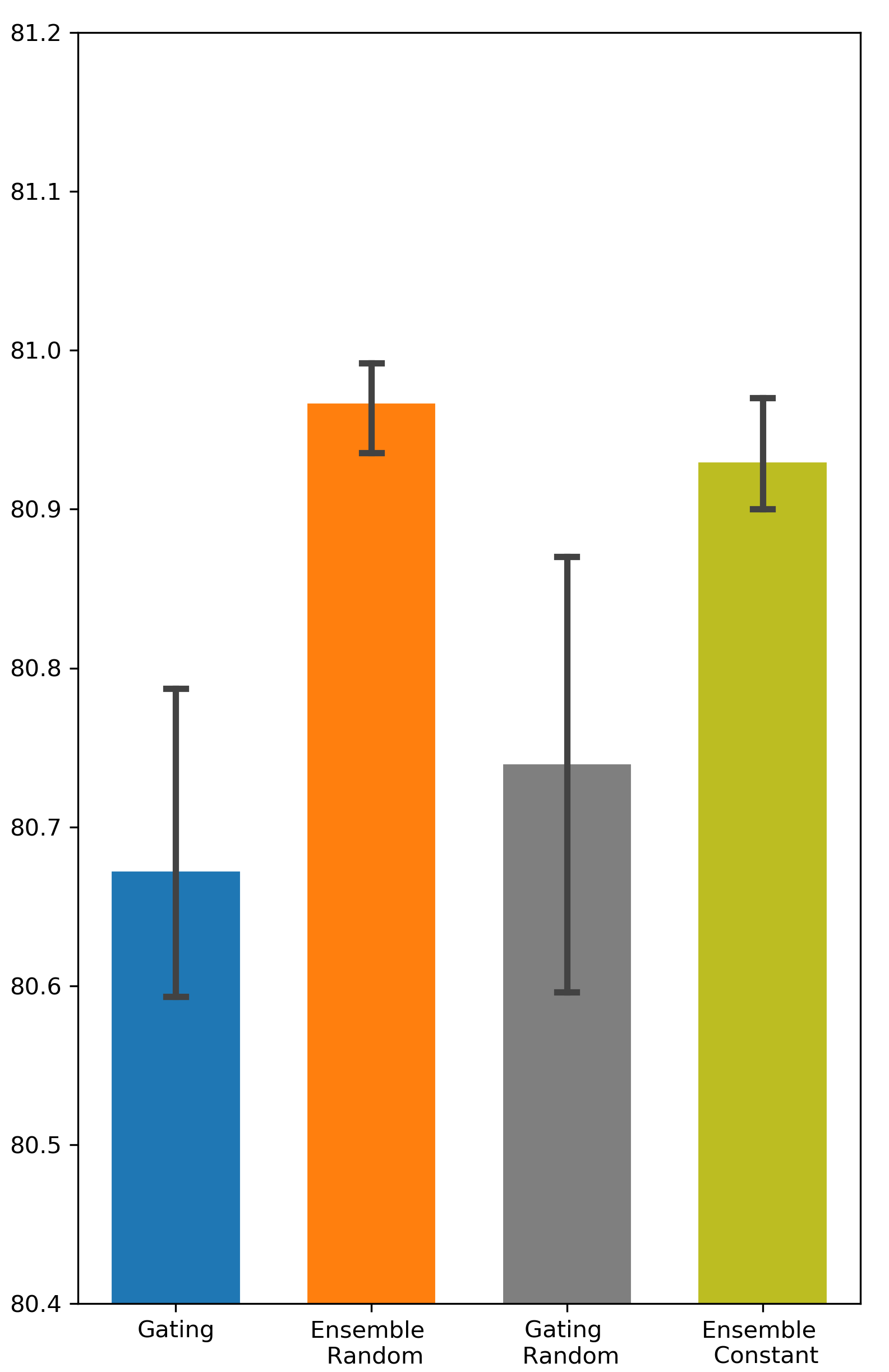}}
\subfloat[DeepFM]{\includegraphics[width=0.3\linewidth]{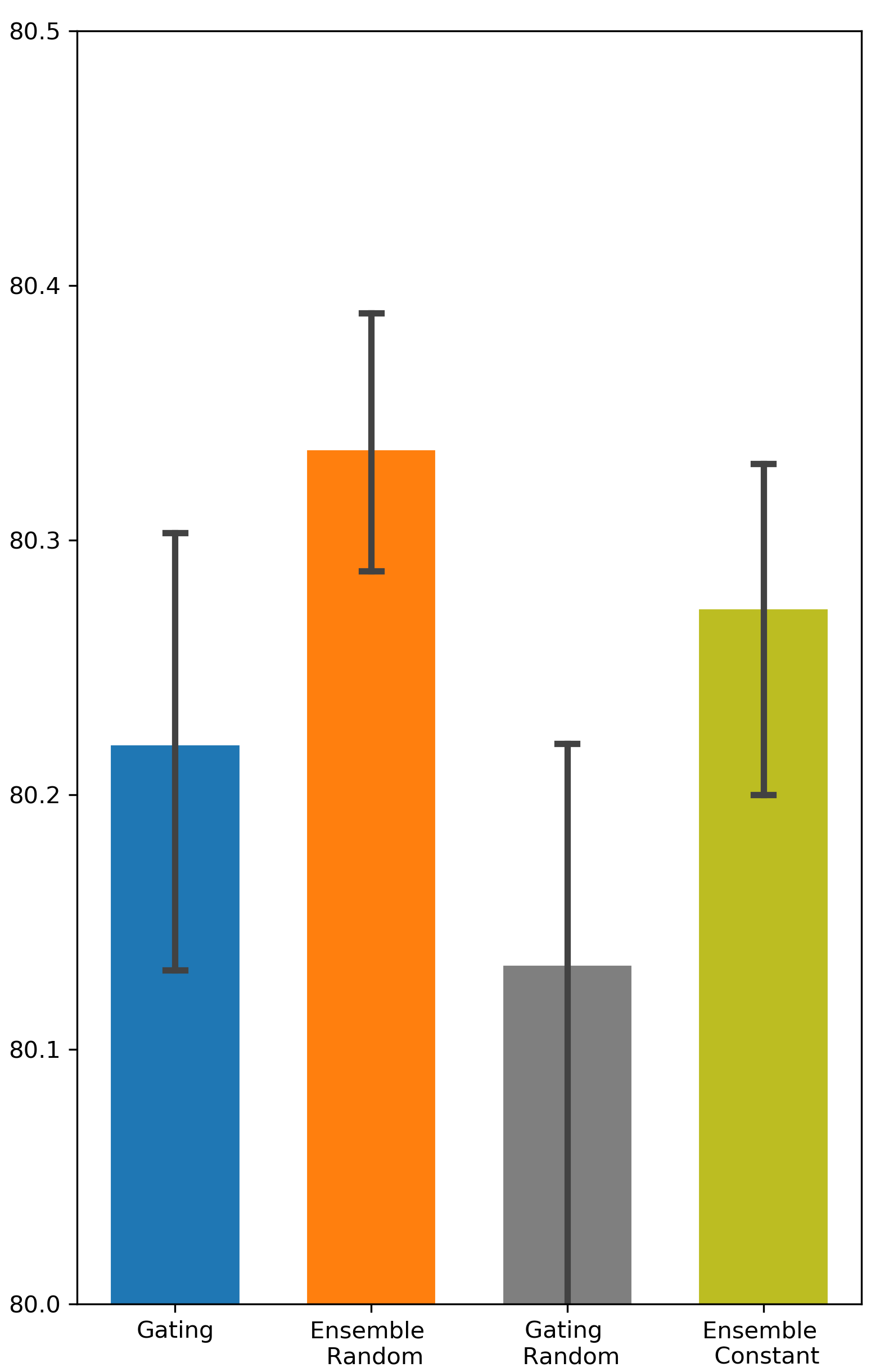}}
\caption{Comparison between random weight initialization and constant weight initialization on Criteo. From left to right, each bar represents Gating (blue), Ensemble Gating (orange), Gating Random (gray), and Ensemble Constant (olive), respectively.}
\label{fig:appendix-init-criteo}
\end{figure}

\begin{figure}
\centering
\subfloat[DCN]{\includegraphics[width=0.3\linewidth]{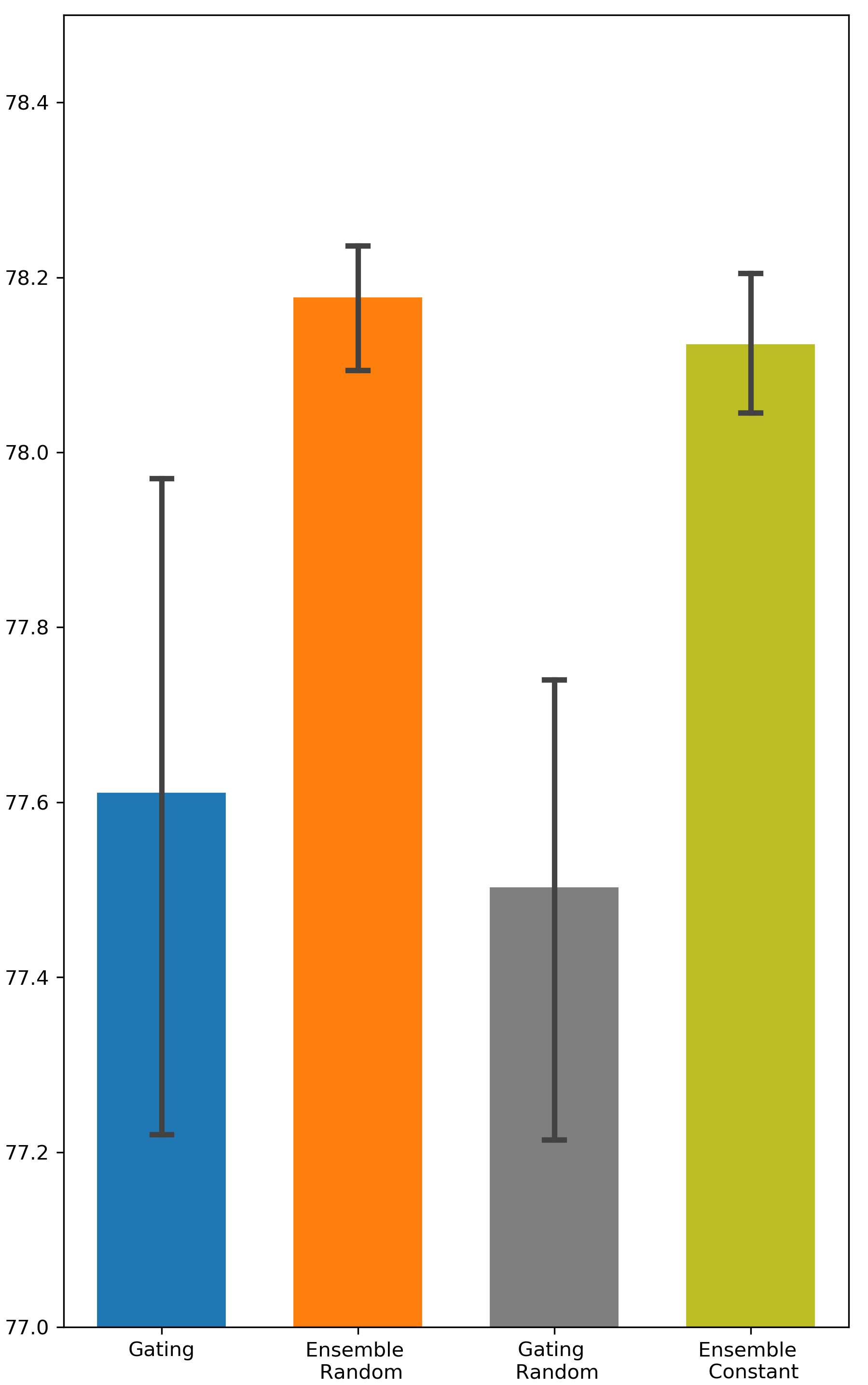}}
\subfloat[AutoInt]{\includegraphics[width=0.3\linewidth]{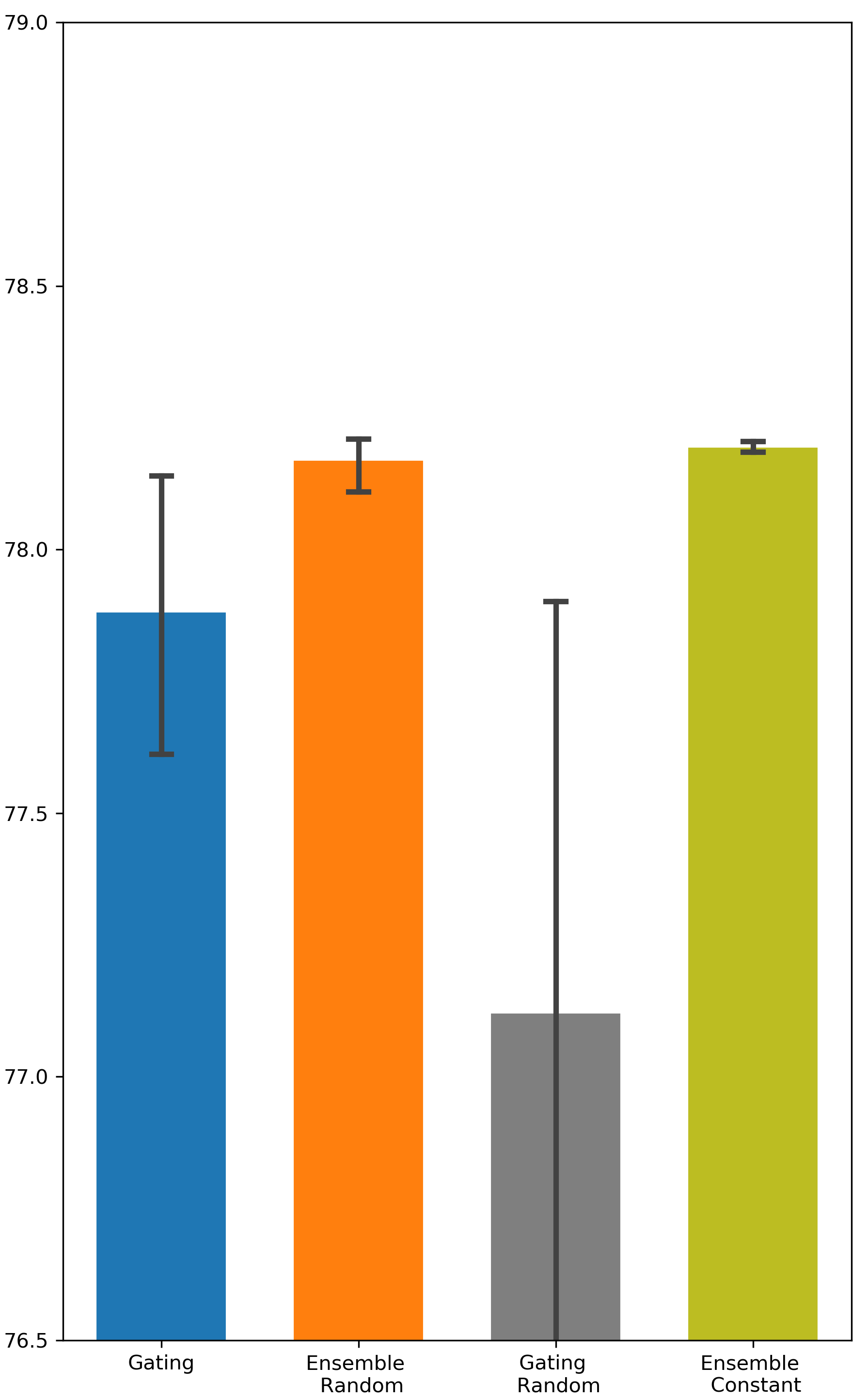}}
\subfloat[DeepFM]{\includegraphics[width=0.3\linewidth]{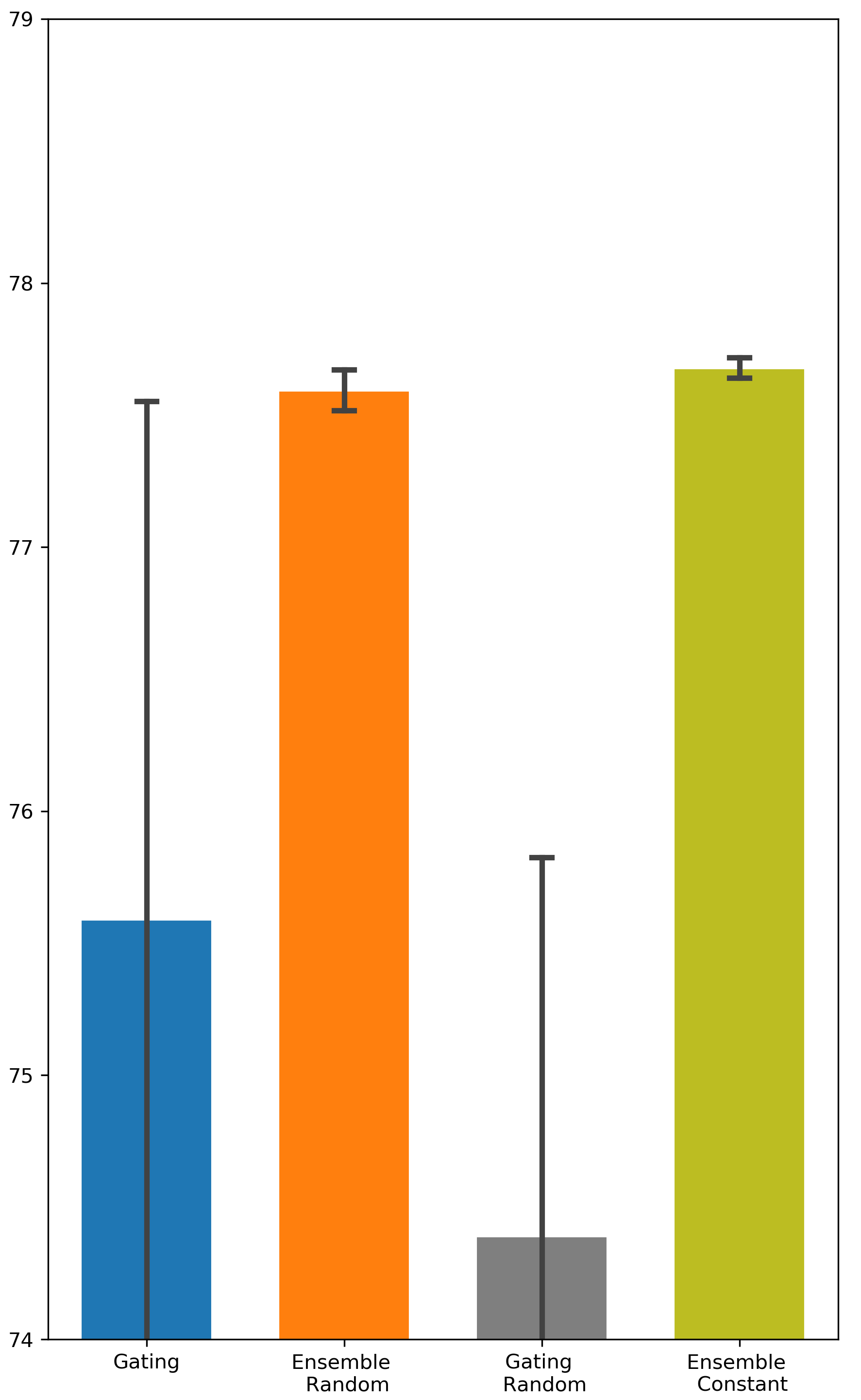}}
\caption{Comparison between random weight initialization and constant weight initialization on Avazu. From left to right, each bar represents Gating (blue), Ensemble Gating (orange), Gating Random (gray), and Ensemble Constant (olive), respectively.}
\label{fig:appendix-init-avazu}
\end{figure}

We used two different methods, i.e. random initialization and constant initialization, to initialize the binary gates. The results can be found in Fig. \ref{fig:appendix-init-criteo} and Fig. \ref{fig:appendix-init-avazu}.

\subsubsection{Comparison between Searching on Validation Set and Searching on Training Set } 
\begin{figure}
\centering
\subfloat[DCN]{\includegraphics[width=0.3\linewidth]{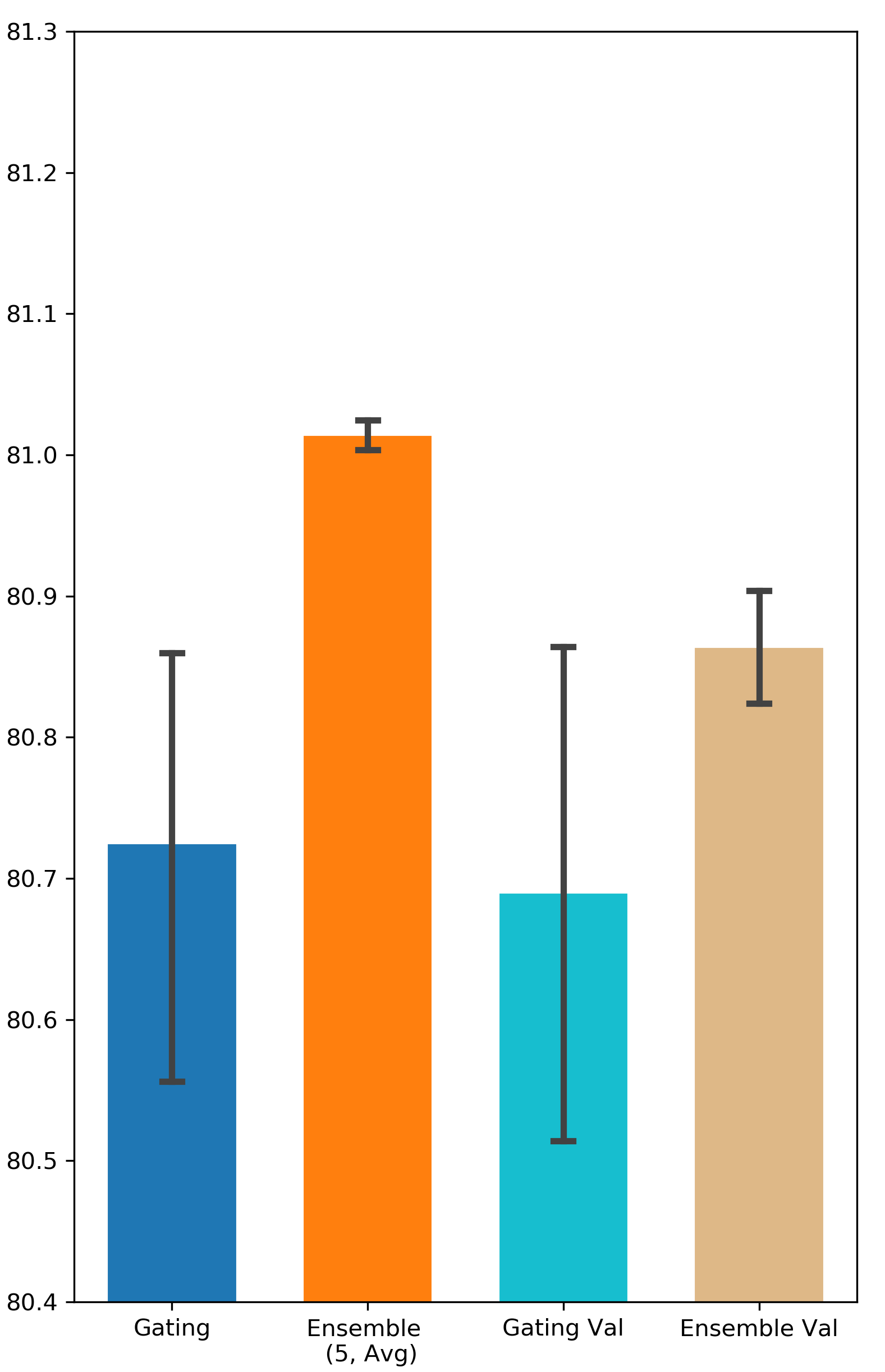}}
\subfloat[AutoInt]{\includegraphics[width=0.3\linewidth]{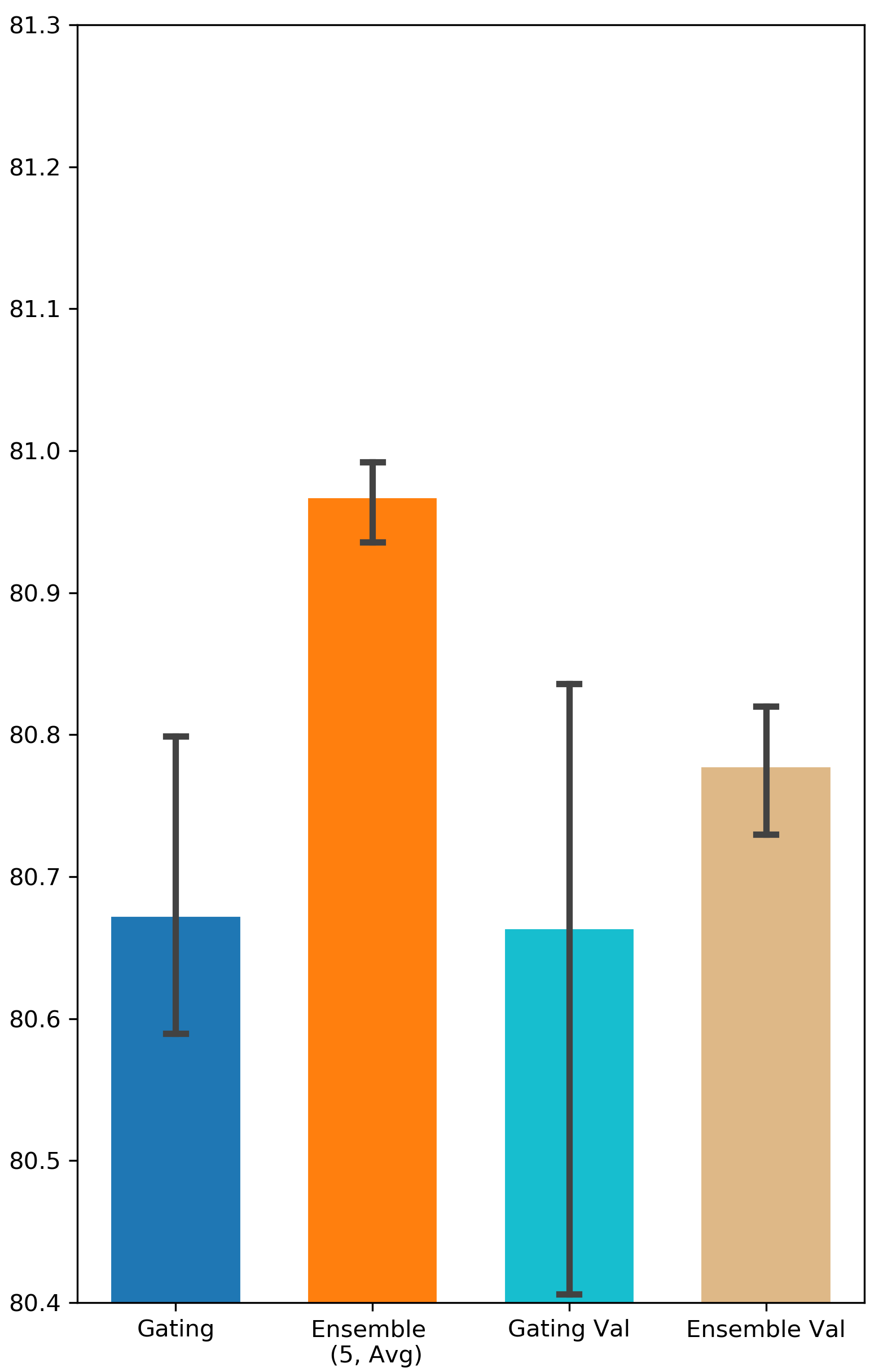}}
\subfloat[DeepFM]{\includegraphics[width=0.3\linewidth]{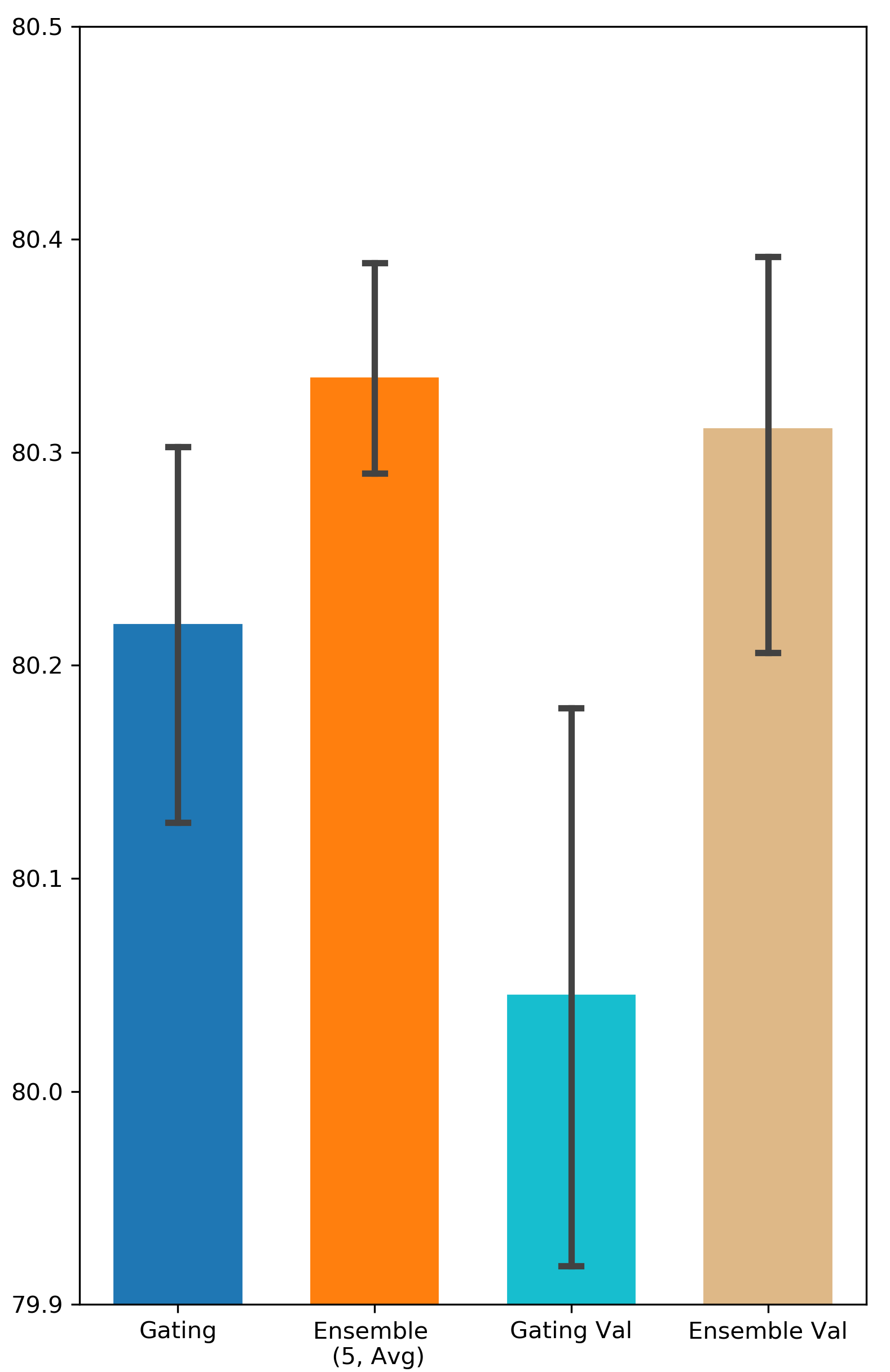}}
\caption{Comparison between searching on validation set and searching on training set. The reported results were obtained on Criteo. From left to right, each bar represents Gating (blue), Ensemble Gating (orange), Gating Val (cyan), and Ensemble Val (burlywood), respectively.}
\label{fig:apendix-val-criteo}
\end{figure}

\begin{figure}
\centering
\subfloat[DCN]{\includegraphics[width=0.3\linewidth]{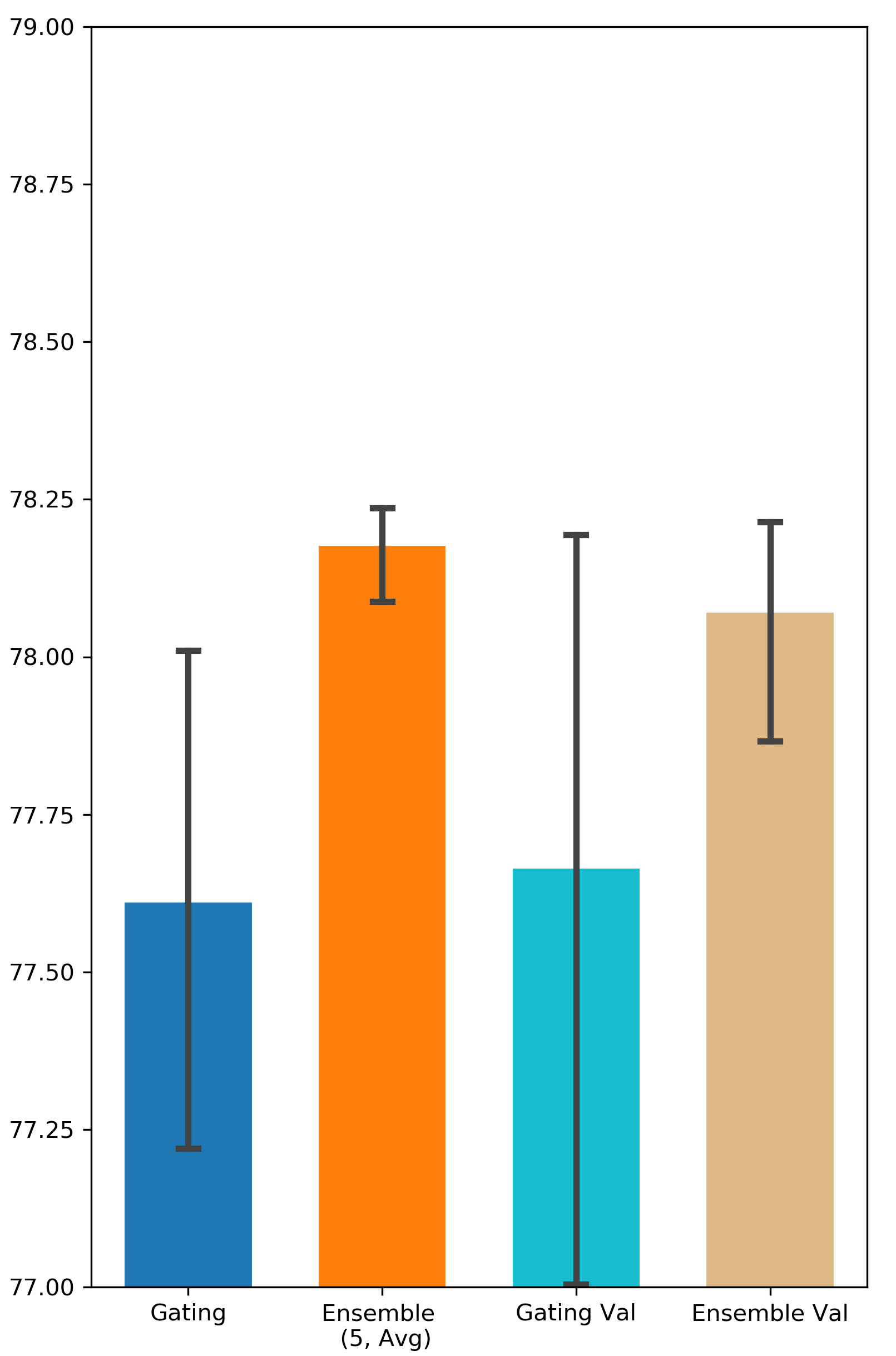}}
\subfloat[AutoInt]{\includegraphics[width=0.3\linewidth]{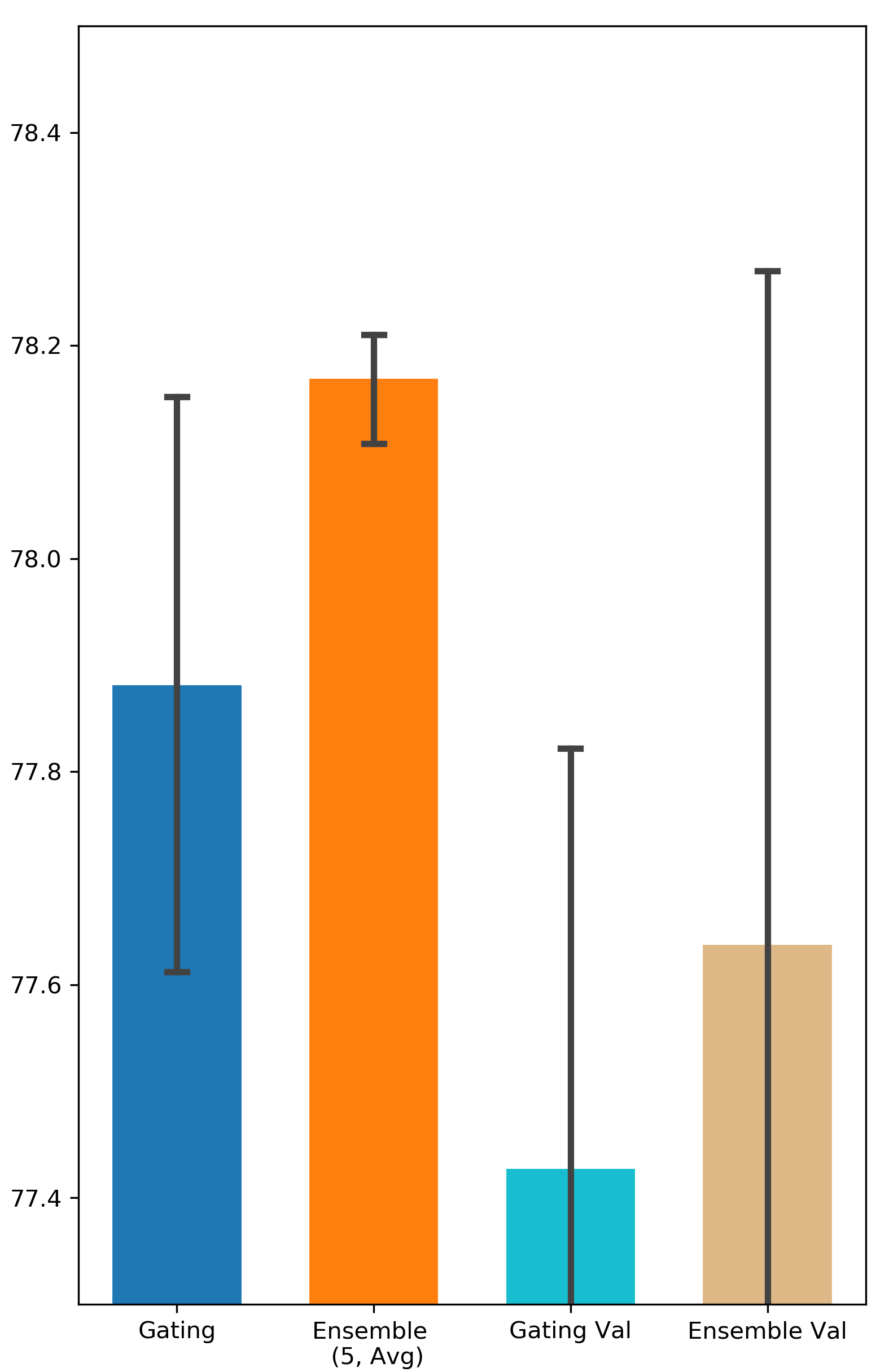}}
\subfloat[DeepFM]{\includegraphics[width=0.3\linewidth]{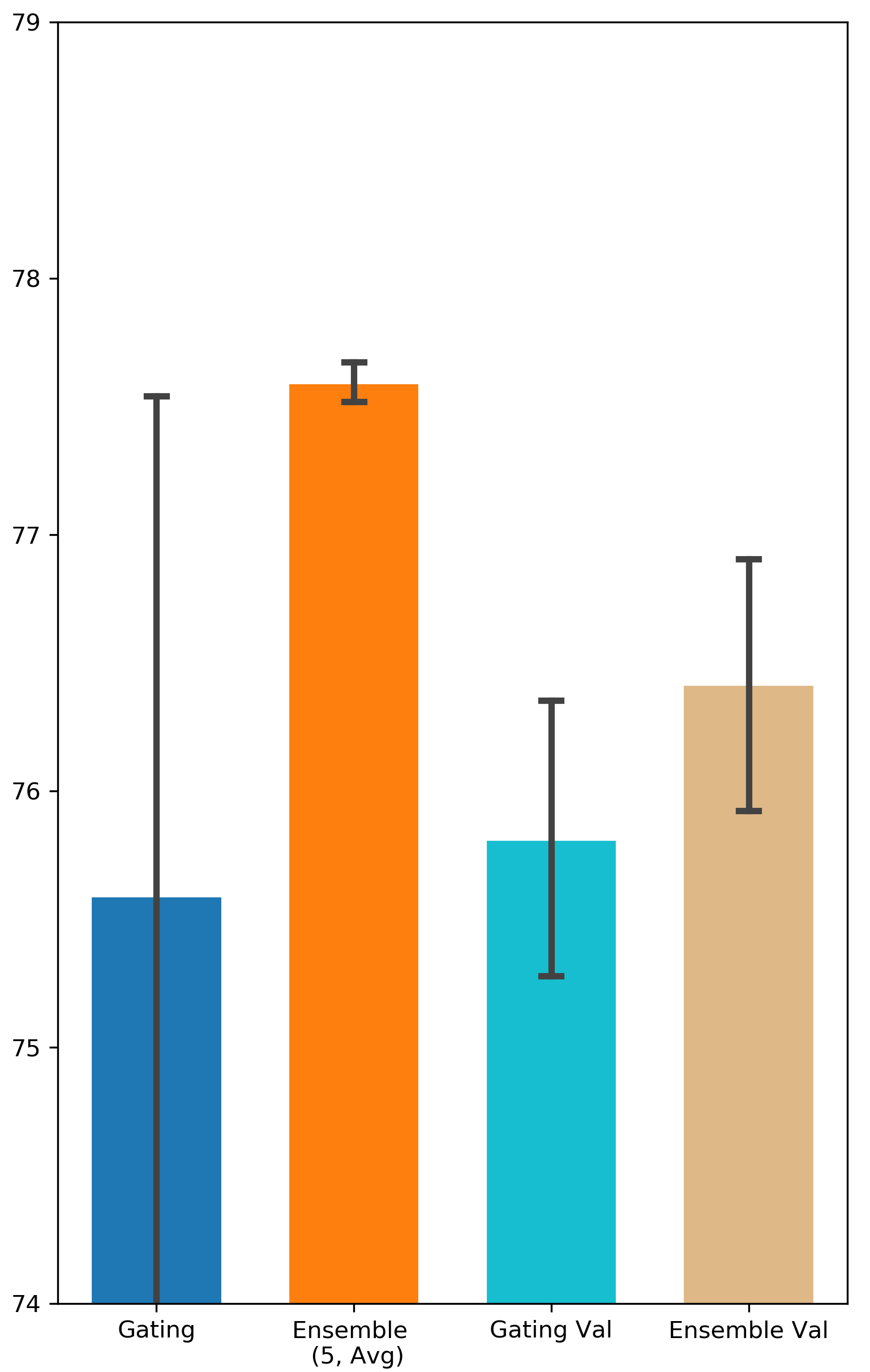}}
\caption{Comparison between searching on validation set and searching on training set. The reported results were obtained on Avazu. From left to right, each bar represents Gating (blue), Ensemble Gating (orange), Gating Val (cyan), and Ensemble Val (burlywood), respectively.}
\label{fig:apendix-val-avazu}
\end{figure}

We intend to investigate the importance of training gating parameters on a separate validation set. We denote the methods that use separate validation set as \textit{Gating Val} and \textit{Ensemble Val}. The results can be found in Fig. \ref{fig:apendix-val-criteo} and Fig. \ref{fig:apendix-val-avazu}.

